\documentclass[10pt,twocolumn,letterpaper]{article}

\usepackage{iccv}
\usepackage{times}
\usepackage{epsfig}
\usepackage{graphicx}
\usepackage{amsmath}
\usepackage{breqn}
\usepackage{amssymb}

\usepackage{booktabs}
\usepackage{color}
\usepackage{comment}
\usepackage{float}
\usepackage{subfigure}
\usepackage{algorithm,algorithmic}
\usepackage{mathtools}
\usepackage{wasysym}
\usepackage[table,xcdraw]{xcolor}
\usepackage{multirow}
\DeclareMathOperator*{\argmin}{argmin}

\definecolor{AliceBlue}{rgb}{0.94,0.97,1.00}
\definecolor{AntiqueWhite1}{rgb}{1.00,0.94,0.86}
\definecolor{AntiqueWhite2}{rgb}{0.93,0.87,0.80}
\definecolor{AntiqueWhite3}{rgb}{0.80,0.75,0.69}
\definecolor{AntiqueWhite4}{rgb}{0.55,0.51,0.47}
\definecolor{AntiqueWhite}{rgb}{0.98,0.92,0.84}
\definecolor{BlanchedAlmond}{rgb}{1.00,0.92,0.80}
\definecolor{BlueViolet}{rgb}{0.54,0.17,0.89}
\definecolor{CadetBlue1}{rgb}{0.60,0.96,1.00}
\definecolor{CadetBlue2}{rgb}{0.56,0.90,0.93}
\definecolor{CadetBlue3}{rgb}{0.48,0.77,0.80}
\definecolor{CadetBlue4}{rgb}{0.33,0.53,0.55}
\definecolor{CadetBlue}{rgb}{0.37,0.62,0.63}
\definecolor{CornflowerBlue}{rgb}{0.39,0.58,0.93}
\definecolor{DarkBlue}{rgb}{0.00,0.00,0.55}
\definecolor{DarkCyan}{rgb}{0.00,0.55,0.55}
\definecolor{DarkGoldenrod1}{rgb}{1.00,0.73,0.06}
\definecolor{DarkGoldenrod2}{rgb}{0.93,0.68,0.05}
\definecolor{DarkGoldenrod3}{rgb}{0.80,0.58,0.05}
\definecolor{DarkGoldenrod4}{rgb}{0.55,0.40,0.03}
\definecolor{DarkGoldenrod}{rgb}{0.72,0.53,0.04}
\definecolor{DarkGray}{rgb}{0.66,0.66,0.66}
\definecolor{DarkGreen}{rgb}{0.00,0.39,0.00}
\definecolor{DarkGrey}{rgb}{0.66,0.66,0.66}
\definecolor{DarkKhaki}{rgb}{0.74,0.72,0.42}
\definecolor{DarkMagenta}{rgb}{0.55,0.00,0.55}
\definecolor{DarkOliveGreen1}{rgb}{0.79,1.00,0.44}
\definecolor{DarkOliveGreen2}{rgb}{0.74,0.93,0.41}
\definecolor{DarkOliveGreen3}{rgb}{0.64,0.80,0.35}
\definecolor{DarkOliveGreen4}{rgb}{0.43,0.55,0.24}
\definecolor{DarkOliveGreen}{rgb}{0.33,0.42,0.18}
\definecolor{DarkOrange1}{rgb}{1.00,0.50,0.00}
\definecolor{DarkOrange2}{rgb}{0.93,0.46,0.00}
\definecolor{DarkOrange3}{rgb}{0.80,0.40,0.00}
\definecolor{DarkOrange4}{rgb}{0.55,0.27,0.00}
\definecolor{DarkOrange}{rgb}{1.00,0.55,0.00}
\definecolor{DarkOrchid1}{rgb}{0.75,0.24,1.00}
\definecolor{DarkOrchid2}{rgb}{0.70,0.23,0.93}
\definecolor{DarkOrchid3}{rgb}{0.60,0.20,0.80}
\definecolor{DarkOrchid4}{rgb}{0.41,0.13,0.55}
\definecolor{DarkOrchid}{rgb}{0.60,0.20,0.80}
\definecolor{DarkRed}{rgb}{0.55,0.00,0.00}
\definecolor{DarkSalmon}{rgb}{0.91,0.59,0.48}
\definecolor{DarkSeaGreen1}{rgb}{0.76,1.00,0.76}
\definecolor{DarkSeaGreen2}{rgb}{0.71,0.93,0.71}
\definecolor{DarkSeaGreen3}{rgb}{0.61,0.80,0.61}
\definecolor{DarkSeaGreen4}{rgb}{0.41,0.55,0.41}
\definecolor{DarkSeaGreen}{rgb}{0.56,0.74,0.56}
\definecolor{DarkSlateBlue}{rgb}{0.28,0.24,0.55}
\definecolor{DarkSlateGray1}{rgb}{0.59,1.00,1.00}
\definecolor{DarkSlateGray2}{rgb}{0.55,0.93,0.93}
\definecolor{DarkSlateGray3}{rgb}{0.47,0.80,0.80}
\definecolor{DarkSlateGray4}{rgb}{0.32,0.55,0.55}
\definecolor{DarkSlateGray}{rgb}{0.18,0.31,0.31}
\definecolor{DarkSlateGrey}{rgb}{0.18,0.31,0.31}
\definecolor{DarkTurquoise}{rgb}{0.00,0.81,0.82}
\definecolor{DarkViolet}{rgb}{0.58,0.00,0.83}
\definecolor{DeepPink1}{rgb}{1.00,0.08,0.58}
\definecolor{DeepPink2}{rgb}{0.93,0.07,0.54}
\definecolor{DeepPink3}{rgb}{0.80,0.06,0.46}
\definecolor{DeepPink4}{rgb}{0.55,0.04,0.31}
\definecolor{DeepPink}{rgb}{1.00,0.08,0.58}
\definecolor{DeepSkyBlue1}{rgb}{0.00,0.75,1.00}
\definecolor{DeepSkyBlue2}{rgb}{0.00,0.70,0.93}
\definecolor{DeepSkyBlue3}{rgb}{0.00,0.60,0.80}
\definecolor{DeepSkyBlue4}{rgb}{0.00,0.41,0.55}
\definecolor{DeepSkyBlue}{rgb}{0.00,0.75,1.00}
\definecolor{DimGray}{rgb}{0.41,0.41,0.41}
\definecolor{DimGrey}{rgb}{0.41,0.41,0.41}
\definecolor{DodgerBlue1}{rgb}{0.12,0.56,1.00}
\definecolor{DodgerBlue2}{rgb}{0.11,0.53,0.93}
\definecolor{DodgerBlue3}{rgb}{0.09,0.45,0.80}
\definecolor{DodgerBlue4}{rgb}{0.06,0.31,0.55}
\definecolor{DodgerBlue}{rgb}{0.12,0.56,1.00}
\definecolor{FloralWhite}{rgb}{1.00,0.98,0.94}
\definecolor{ForestGreen}{rgb}{0.13,0.55,0.13}
\definecolor{GhostWhite}{rgb}{0.97,0.97,1.00}
\definecolor{GreenYellow}{rgb}{0.68,1.00,0.18}
\definecolor{HotPink1}{rgb}{1.00,0.43,0.71}
\definecolor{HotPink2}{rgb}{0.93,0.42,0.65}
\definecolor{HotPink3}{rgb}{0.80,0.38,0.56}
\definecolor{HotPink4}{rgb}{0.55,0.23,0.38}
\definecolor{HotPink}{rgb}{1.00,0.41,0.71}
\definecolor{IndianRed1}{rgb}{1.00,0.42,0.42}
\definecolor{IndianRed2}{rgb}{0.93,0.39,0.39}
\definecolor{IndianRed3}{rgb}{0.80,0.33,0.33}
\definecolor{IndianRed4}{rgb}{0.55,0.23,0.23}
\definecolor{IndianRed}{rgb}{0.80,0.36,0.36}
\definecolor{LavenderBlush1}{rgb}{1.00,0.94,0.96}
\definecolor{LavenderBlush2}{rgb}{0.93,0.88,0.90}
\definecolor{LavenderBlush3}{rgb}{0.80,0.76,0.77}
\definecolor{LavenderBlush4}{rgb}{0.55,0.51,0.53}
\definecolor{LavenderBlush}{rgb}{1.00,0.94,0.96}
\definecolor{LawnGreen}{rgb}{0.49,0.99,0.00}
\definecolor{LemonChiffon1}{rgb}{1.00,0.98,0.80}
\definecolor{LemonChiffon2}{rgb}{0.93,0.91,0.75}
\definecolor{LemonChiffon3}{rgb}{0.80,0.79,0.65}
\definecolor{LemonChiffon4}{rgb}{0.55,0.54,0.44}
\definecolor{LemonChiffon}{rgb}{1.00,0.98,0.80}
\definecolor{LightBlue1}{rgb}{0.75,0.94,1.00}
\definecolor{LightBlue2}{rgb}{0.70,0.87,0.93}
\definecolor{LightBlue3}{rgb}{0.60,0.75,0.80}
\definecolor{LightBlue4}{rgb}{0.41,0.51,0.55}
\definecolor{LightBlue}{rgb}{0.68,0.85,0.90}
\definecolor{LightCoral}{rgb}{0.94,0.50,0.50}
\definecolor{LightCyan1}{rgb}{0.88,1.00,1.00}
\definecolor{LightCyan2}{rgb}{0.82,0.93,0.93}
\definecolor{LightCyan3}{rgb}{0.71,0.80,0.80}
\definecolor{LightCyan4}{rgb}{0.48,0.55,0.55}
\definecolor{LightCyan}{rgb}{0.88,1.00,1.00}
\definecolor{LightGoldenrod1}{rgb}{1.00,0.93,0.55}
\definecolor{LightGoldenrod2}{rgb}{0.93,0.86,0.51}
\definecolor{LightGoldenrod3}{rgb}{0.80,0.75,0.44}
\definecolor{LightGoldenrod4}{rgb}{0.55,0.51,0.30}
\definecolor{LightGoldenrodYellow}{rgb}{0.98,0.98,0.82}
\definecolor{LightGoldenrod}{rgb}{0.93,0.87,0.51}
\definecolor{LightGray}{rgb}{0.83,0.83,0.83}
\definecolor{LightGreen}{rgb}{0.56,0.93,0.56}
\definecolor{LightGrey}{rgb}{0.83,0.83,0.83}
\definecolor{LightPink1}{rgb}{1.00,0.68,0.73}
\definecolor{LightPink2}{rgb}{0.93,0.64,0.68}
\definecolor{LightPink3}{rgb}{0.80,0.55,0.58}
\definecolor{LightPink4}{rgb}{0.55,0.37,0.40}
\definecolor{LightPink}{rgb}{1.00,0.71,0.76}
\definecolor{LightSalmon1}{rgb}{1.00,0.63,0.48}
\definecolor{LightSalmon2}{rgb}{0.93,0.58,0.45}
\definecolor{LightSalmon3}{rgb}{0.80,0.51,0.38}
\definecolor{LightSalmon4}{rgb}{0.55,0.34,0.26}
\definecolor{LightSalmon}{rgb}{1.00,0.63,0.48}
\definecolor{LightSeaGreen}{rgb}{0.13,0.70,0.67}
\definecolor{LightSkyBlue1}{rgb}{0.69,0.89,1.00}
\definecolor{LightSkyBlue2}{rgb}{0.64,0.83,0.93}
\definecolor{LightSkyBlue3}{rgb}{0.55,0.71,0.80}
\definecolor{LightSkyBlue4}{rgb}{0.38,0.48,0.55}
\definecolor{LightSkyBlue}{rgb}{0.53,0.81,0.98}
\definecolor{LightSlateBlue}{rgb}{0.52,0.44,1.00}
\definecolor{LightSlateGray}{rgb}{0.47,0.53,0.60}
\definecolor{LightSlateGrey}{rgb}{0.47,0.53,0.60}
\definecolor{LightSteelBlue1}{rgb}{0.79,0.88,1.00}
\definecolor{LightSteelBlue2}{rgb}{0.74,0.82,0.93}
\definecolor{LightSteelBlue3}{rgb}{0.64,0.71,0.80}
\definecolor{LightSteelBlue4}{rgb}{0.43,0.48,0.55}
\definecolor{LightSteelBlue}{rgb}{0.69,0.77,0.87}
\definecolor{LightYellow1}{rgb}{1.00,1.00,0.88}
\definecolor{LightYellow2}{rgb}{0.93,0.93,0.82}
\definecolor{LightYellow3}{rgb}{0.80,0.80,0.71}
\definecolor{LightYellow4}{rgb}{0.55,0.55,0.48}
\definecolor{LightYellow}{rgb}{1.00,1.00,0.88}
\definecolor{LimeGreen}{rgb}{0.20,0.80,0.20}
\definecolor{MediumAquamarine}{rgb}{0.40,0.80,0.67}
\definecolor{MediumBlue}{rgb}{0.00,0.00,0.80}
\definecolor{MediumOrchid1}{rgb}{0.88,0.40,1.00}
\definecolor{MediumOrchid2}{rgb}{0.82,0.37,0.93}
\definecolor{MediumOrchid3}{rgb}{0.71,0.32,0.80}
\definecolor{MediumOrchid4}{rgb}{0.48,0.22,0.55}
\definecolor{MediumOrchid}{rgb}{0.73,0.33,0.83}
\definecolor{MediumPurple1}{rgb}{0.67,0.51,1.00}
\definecolor{MediumPurple2}{rgb}{0.62,0.47,0.93}
\definecolor{MediumPurple3}{rgb}{0.54,0.41,0.80}
\definecolor{MediumPurple4}{rgb}{0.36,0.28,0.55}
\definecolor{MediumPurple}{rgb}{0.58,0.44,0.86}
\definecolor{MediumSeaGreen}{rgb}{0.24,0.70,0.44}
\definecolor{MediumSlateBlue}{rgb}{0.48,0.41,0.93}
\definecolor{MediumSpringGreen}{rgb}{0.00,0.98,0.60}
\definecolor{MediumTurquoise}{rgb}{0.28,0.82,0.80}
\definecolor{MediumVioletRed}{rgb}{0.78,0.08,0.52}
\definecolor{MidnightBlue}{rgb}{0.10,0.10,0.44}
\definecolor{MintCream}{rgb}{0.96,1.00,0.98}
\definecolor{MistyRose1}{rgb}{1.00,0.89,0.88}
\definecolor{MistyRose2}{rgb}{0.93,0.84,0.82}
\definecolor{MistyRose3}{rgb}{0.80,0.72,0.71}
\definecolor{MistyRose4}{rgb}{0.55,0.49,0.48}
\definecolor{MistyRose}{rgb}{1.00,0.89,0.88}
\definecolor{NavajoWhite1}{rgb}{1.00,0.87,0.68}
\definecolor{NavajoWhite2}{rgb}{0.93,0.81,0.63}
\definecolor{NavajoWhite3}{rgb}{0.80,0.70,0.55}
\definecolor{NavajoWhite4}{rgb}{0.55,0.47,0.37}
\definecolor{NavajoWhite}{rgb}{1.00,0.87,0.68}
\definecolor{NavyBlue}{rgb}{0.00,0.00,0.50}
\definecolor{OldLace}{rgb}{0.99,0.96,0.90}
\definecolor{OliveDrab1}{rgb}{0.75,1.00,0.24}
\definecolor{OliveDrab2}{rgb}{0.70,0.93,0.23}
\definecolor{OliveDrab3}{rgb}{0.60,0.80,0.20}
\definecolor{OliveDrab4}{rgb}{0.41,0.55,0.13}
\definecolor{OliveDrab}{rgb}{0.42,0.56,0.14}
\definecolor{OrangeRed1}{rgb}{1.00,0.27,0.00}
\definecolor{OrangeRed2}{rgb}{0.93,0.25,0.00}
\definecolor{OrangeRed3}{rgb}{0.80,0.22,0.00}
\definecolor{OrangeRed4}{rgb}{0.55,0.15,0.00}
\definecolor{OrangeRed}{rgb}{1.00,0.27,0.00}
\definecolor{PaleGoldenrod}{rgb}{0.93,0.91,0.67}
\definecolor{PaleGreen1}{rgb}{0.60,1.00,0.60}
\definecolor{PaleGreen2}{rgb}{0.56,0.93,0.56}
\definecolor{PaleGreen3}{rgb}{0.49,0.80,0.49}
\definecolor{PaleGreen4}{rgb}{0.33,0.55,0.33}
\definecolor{PaleGreen}{rgb}{0.60,0.98,0.60}
\definecolor{PaleTurquoise1}{rgb}{0.73,1.00,1.00}
\definecolor{PaleTurquoise2}{rgb}{0.68,0.93,0.93}
\definecolor{PaleTurquoise3}{rgb}{0.59,0.80,0.80}
\definecolor{PaleTurquoise4}{rgb}{0.40,0.55,0.55}
\definecolor{PaleTurquoise}{rgb}{0.69,0.93,0.93}
\definecolor{PaleVioletRed1}{rgb}{1.00,0.51,0.67}
\definecolor{PaleVioletRed2}{rgb}{0.93,0.47,0.62}
\definecolor{PaleVioletRed3}{rgb}{0.80,0.41,0.54}
\definecolor{PaleVioletRed4}{rgb}{0.55,0.28,0.36}
\definecolor{PaleVioletRed}{rgb}{0.86,0.44,0.58}
\definecolor{PapayaWhip}{rgb}{1.00,0.94,0.84}
\definecolor{PeachPuff1}{rgb}{1.00,0.85,0.73}
\definecolor{PeachPuff2}{rgb}{0.93,0.80,0.68}
\definecolor{PeachPuff3}{rgb}{0.80,0.69,0.58}
\definecolor{PeachPuff4}{rgb}{0.55,0.47,0.40}
\definecolor{PeachPuff}{rgb}{1.00,0.85,0.73}
\definecolor{PowderBlue}{rgb}{0.69,0.88,0.90}
\definecolor{RosyBrown1}{rgb}{1.00,0.76,0.76}
\definecolor{RosyBrown2}{rgb}{0.93,0.71,0.71}
\definecolor{RosyBrown3}{rgb}{0.80,0.61,0.61}
\definecolor{RosyBrown4}{rgb}{0.55,0.41,0.41}
\definecolor{RosyBrown}{rgb}{0.74,0.56,0.56}
\definecolor{RoyalBlue1}{rgb}{0.28,0.46,1.00}
\definecolor{RoyalBlue2}{rgb}{0.26,0.43,0.93}
\definecolor{RoyalBlue3}{rgb}{0.23,0.37,0.80}
\definecolor{RoyalBlue4}{rgb}{0.15,0.25,0.55}
\definecolor{RoyalBlue}{rgb}{0.25,0.41,0.88}
\definecolor{SaddleBrown}{rgb}{0.55,0.27,0.07}
\definecolor{SandyBrown}{rgb}{0.96,0.64,0.38}
\definecolor{SeaGreen1}{rgb}{0.33,1.00,0.62}
\definecolor{SeaGreen2}{rgb}{0.31,0.93,0.58}
\definecolor{SeaGreen3}{rgb}{0.26,0.80,0.50}
\definecolor{SeaGreen4}{rgb}{0.18,0.55,0.34}
\definecolor{SeaGreen}{rgb}{0.18,0.55,0.34}
\definecolor{SkyBlue1}{rgb}{0.53,0.81,1.00}
\definecolor{SkyBlue2}{rgb}{0.49,0.75,0.93}
\definecolor{SkyBlue3}{rgb}{0.42,0.65,0.80}
\definecolor{SkyBlue4}{rgb}{0.29,0.44,0.55}
\definecolor{SkyBlue}{rgb}{0.53,0.81,0.92}
\definecolor{SlateBlue1}{rgb}{0.51,0.44,1.00}
\definecolor{SlateBlue2}{rgb}{0.48,0.40,0.93}
\definecolor{SlateBlue3}{rgb}{0.41,0.35,0.80}
\definecolor{SlateBlue4}{rgb}{0.28,0.24,0.55}
\definecolor{SlateBlue}{rgb}{0.42,0.35,0.80}
\definecolor{SlateGray1}{rgb}{0.78,0.89,1.00}
\definecolor{SlateGray2}{rgb}{0.73,0.83,0.93}
\definecolor{SlateGray3}{rgb}{0.62,0.71,0.80}
\definecolor{SlateGray4}{rgb}{0.42,0.48,0.55}
\definecolor{SlateGray}{rgb}{0.44,0.50,0.56}
\definecolor{SlateGrey}{rgb}{0.44,0.50,0.56}
\definecolor{SpringGreen1}{rgb}{0.00,1.00,0.50}
\definecolor{SpringGreen2}{rgb}{0.00,0.93,0.46}
\definecolor{SpringGreen3}{rgb}{0.00,0.80,0.40}
\definecolor{SpringGreen4}{rgb}{0.00,0.55,0.27}
\definecolor{SpringGreen}{rgb}{0.00,1.00,0.50}
\definecolor{SteelBlue1}{rgb}{0.39,0.72,1.00}
\definecolor{SteelBlue2}{rgb}{0.36,0.67,0.93}
\definecolor{SteelBlue3}{rgb}{0.31,0.58,0.80}
\definecolor{SteelBlue4}{rgb}{0.21,0.39,0.55}
\definecolor{SteelBlue}{rgb}{0.27,0.51,0.71}
\definecolor{VioletRed1}{rgb}{1.00,0.24,0.59}
\definecolor{VioletRed2}{rgb}{0.93,0.23,0.55}
\definecolor{VioletRed3}{rgb}{0.80,0.20,0.47}
\definecolor{VioletRed4}{rgb}{0.55,0.13,0.32}
\definecolor{VioletRed}{rgb}{0.82,0.13,0.56}
\definecolor{WhiteSmoke}{rgb}{0.96,0.96,0.96}
\definecolor{YellowGreen}{rgb}{0.60,0.80,0.20}
\definecolor{aliceblue}{rgb}{0.94,0.97,1.00}
\definecolor{antiquewhite}{rgb}{0.98,0.92,0.84}
\definecolor{aquamarine1}{rgb}{0.50,1.00,0.83}
\definecolor{aquamarine2}{rgb}{0.46,0.93,0.78}
\definecolor{aquamarine3}{rgb}{0.40,0.80,0.67}
\definecolor{aquamarine4}{rgb}{0.27,0.55,0.45}
\definecolor{aquamarine}{rgb}{0.50,1.00,0.83}
\definecolor{azure1}{rgb}{0.94,1.00,1.00}
\definecolor{azure2}{rgb}{0.88,0.93,0.93}
\definecolor{azure3}{rgb}{0.76,0.80,0.80}
\definecolor{azure4}{rgb}{0.51,0.55,0.55}
\definecolor{azure}{rgb}{0.94,1.00,1.00}
\definecolor{beige}{rgb}{0.96,0.96,0.86}
\definecolor{bisque1}{rgb}{1.00,0.89,0.77}
\definecolor{bisque2}{rgb}{0.93,0.84,0.72}
\definecolor{bisque3}{rgb}{0.80,0.72,0.62}
\definecolor{bisque4}{rgb}{0.55,0.49,0.42}
\definecolor{bisque}{rgb}{1.00,0.89,0.77}
\definecolor{black}{rgb}{0.00,0.00,0.00}
\definecolor{blanchedalmond}{rgb}{1.00,0.92,0.80}
\definecolor{blue1}{rgb}{0.00,0.00,1.00}
\definecolor{blue2}{rgb}{0.00,0.00,0.93}
\definecolor{blue3}{rgb}{0.00,0.00,0.80}
\definecolor{blue4}{rgb}{0.00,0.00,0.55}
\definecolor{blueviolet}{rgb}{0.54,0.17,0.89}
\definecolor{blue}{rgb}{0.00,0.00,1.00}
\definecolor{brown1}{rgb}{1.00,0.25,0.25}
\definecolor{brown2}{rgb}{0.93,0.23,0.23}
\definecolor{brown3}{rgb}{0.80,0.20,0.20}
\definecolor{brown4}{rgb}{0.55,0.14,0.14}
\definecolor{brown}{rgb}{0.65,0.16,0.16}
\definecolor{burlywood1}{rgb}{1.00,0.83,0.61}
\definecolor{burlywood2}{rgb}{0.93,0.77,0.57}
\definecolor{burlywood3}{rgb}{0.80,0.67,0.49}
\definecolor{burlywood4}{rgb}{0.55,0.45,0.33}
\definecolor{burlywood}{rgb}{0.87,0.72,0.53}
\definecolor{cadetblue}{rgb}{0.37,0.62,0.63}
\definecolor{chartreuse1}{rgb}{0.50,1.00,0.00}
\definecolor{chartreuse2}{rgb}{0.46,0.93,0.00}
\definecolor{chartreuse3}{rgb}{0.40,0.80,0.00}
\definecolor{chartreuse4}{rgb}{0.27,0.55,0.00}
\definecolor{chartreuse}{rgb}{0.50,1.00,0.00}
\definecolor{chocolate1}{rgb}{1.00,0.50,0.14}
\definecolor{chocolate2}{rgb}{0.93,0.46,0.13}
\definecolor{chocolate3}{rgb}{0.80,0.40,0.11}
\definecolor{chocolate4}{rgb}{0.55,0.27,0.07}
\definecolor{chocolate}{rgb}{0.82,0.41,0.12}
\definecolor{coral1}{rgb}{1.00,0.45,0.34}
\definecolor{coral2}{rgb}{0.93,0.42,0.31}
\definecolor{coral3}{rgb}{0.80,0.36,0.27}
\definecolor{coral4}{rgb}{0.55,0.24,0.18}
\definecolor{coral}{rgb}{1.00,0.50,0.31}
\definecolor{cornflowerblue}{rgb}{0.39,0.58,0.93}
\definecolor{cornsilk1}{rgb}{1.00,0.97,0.86}
\definecolor{cornsilk2}{rgb}{0.93,0.91,0.80}
\definecolor{cornsilk3}{rgb}{0.80,0.78,0.69}
\definecolor{cornsilk4}{rgb}{0.55,0.53,0.47}
\definecolor{cornsilk}{rgb}{1.00,0.97,0.86}
\definecolor{cyan1}{rgb}{0.00,1.00,1.00}
\definecolor{cyan2}{rgb}{0.00,0.93,0.93}
\definecolor{cyan3}{rgb}{0.00,0.80,0.80}
\definecolor{cyan4}{rgb}{0.00,0.55,0.55}
\definecolor{cyan}{rgb}{0.00,1.00,1.00}
\definecolor{darkblue}{rgb}{0.00,0.00,0.55}
\definecolor{darkcyan}{rgb}{0.00,0.55,0.55}
\definecolor{darkgoldenrod}{rgb}{0.72,0.53,0.04}
\definecolor{darkgray}{rgb}{0.66,0.66,0.66}
\definecolor{darkgreen}{rgb}{0.00,0.39,0.00}
\definecolor{darkgrey}{rgb}{0.66,0.66,0.66}
\definecolor{darkkhaki}{rgb}{0.74,0.72,0.42}
\definecolor{darkmagenta}{rgb}{0.55,0.00,0.55}
\definecolor{darkolive}{rgb}{0.33,0.42,0.18}
\definecolor{darkorange}{rgb}{1.00,0.55,0.00}
\definecolor{darkorchid}{rgb}{0.60,0.20,0.80}
\definecolor{darkred}{rgb}{0.55,0.00,0.00}
\definecolor{darksalmon}{rgb}{0.91,0.59,0.48}
\definecolor{darksea}{rgb}{0.56,0.74,0.56}
\definecolor{darkslate}{rgb}{0.18,0.31,0.31}
\definecolor{darkslate}{rgb}{0.18,0.31,0.31}
\definecolor{darkslate}{rgb}{0.28,0.24,0.55}
\definecolor{darkturquoise}{rgb}{0.00,0.81,0.82}
\definecolor{darkviolet}{rgb}{0.58,0.00,0.83}
\definecolor{deeppink}{rgb}{1.00,0.08,0.58}
\definecolor{deepsky}{rgb}{0.00,0.75,1.00}
\definecolor{dimgray}{rgb}{0.41,0.41,0.41}
\definecolor{dimgrey}{rgb}{0.41,0.41,0.41}
\definecolor{dodgerblue}{rgb}{0.12,0.56,1.00}
\definecolor{firebrick1}{rgb}{1.00,0.19,0.19}
\definecolor{firebrick2}{rgb}{0.93,0.17,0.17}
\definecolor{firebrick3}{rgb}{0.80,0.15,0.15}
\definecolor{firebrick4}{rgb}{0.55,0.10,0.10}
\definecolor{firebrick}{rgb}{0.70,0.13,0.13}
\definecolor{floralwhite}{rgb}{1.00,0.98,0.94}
\definecolor{forestgreen}{rgb}{0.13,0.55,0.13}
\definecolor{gainsboro}{rgb}{0.86,0.86,0.86}
\definecolor{ghostwhite}{rgb}{0.97,0.97,1.00}
\definecolor{gold1}{rgb}{1.00,0.84,0.00}
\definecolor{gold2}{rgb}{0.93,0.79,0.00}
\definecolor{gold3}{rgb}{0.80,0.68,0.00}
\definecolor{gold4}{rgb}{0.55,0.46,0.00}
\definecolor{goldenrod1}{rgb}{1.00,0.76,0.15}
\definecolor{goldenrod2}{rgb}{0.93,0.71,0.13}
\definecolor{goldenrod3}{rgb}{0.80,0.61,0.11}
\definecolor{goldenrod4}{rgb}{0.55,0.41,0.08}
\definecolor{goldenrod}{rgb}{0.85,0.65,0.13}
\definecolor{gold}{rgb}{1.00,0.84,0.00}
\definecolor{gray0}{rgb}{0.00,0.00,0.00}
\definecolor{gray100}{rgb}{1.00,1.00,1.00}
\definecolor{gray10}{rgb}{0.10,0.10,0.10}
\definecolor{gray11}{rgb}{0.11,0.11,0.11}
\definecolor{gray12}{rgb}{0.12,0.12,0.12}
\definecolor{gray13}{rgb}{0.13,0.13,0.13}
\definecolor{gray14}{rgb}{0.14,0.14,0.14}
\definecolor{gray15}{rgb}{0.15,0.15,0.15}
\definecolor{gray16}{rgb}{0.16,0.16,0.16}
\definecolor{gray17}{rgb}{0.17,0.17,0.17}
\definecolor{gray18}{rgb}{0.18,0.18,0.18}
\definecolor{gray19}{rgb}{0.19,0.19,0.19}
\definecolor{gray1}{rgb}{0.01,0.01,0.01}
\definecolor{gray20}{rgb}{0.20,0.20,0.20}
\definecolor{gray21}{rgb}{0.21,0.21,0.21}
\definecolor{gray22}{rgb}{0.22,0.22,0.22}
\definecolor{gray23}{rgb}{0.23,0.23,0.23}
\definecolor{gray24}{rgb}{0.24,0.24,0.24}
\definecolor{gray25}{rgb}{0.25,0.25,0.25}
\definecolor{gray26}{rgb}{0.26,0.26,0.26}
\definecolor{gray27}{rgb}{0.27,0.27,0.27}
\definecolor{gray28}{rgb}{0.28,0.28,0.28}
\definecolor{gray29}{rgb}{0.29,0.29,0.29}
\definecolor{gray2}{rgb}{0.02,0.02,0.02}
\definecolor{gray30}{rgb}{0.30,0.30,0.30}
\definecolor{gray31}{rgb}{0.31,0.31,0.31}
\definecolor{gray32}{rgb}{0.32,0.32,0.32}
\definecolor{gray33}{rgb}{0.33,0.33,0.33}
\definecolor{gray34}{rgb}{0.34,0.34,0.34}
\definecolor{gray35}{rgb}{0.35,0.35,0.35}
\definecolor{gray36}{rgb}{0.36,0.36,0.36}
\definecolor{gray37}{rgb}{0.37,0.37,0.37}
\definecolor{gray38}{rgb}{0.38,0.38,0.38}
\definecolor{gray39}{rgb}{0.39,0.39,0.39}
\definecolor{gray3}{rgb}{0.03,0.03,0.03}
\definecolor{gray40}{rgb}{0.40,0.40,0.40}
\definecolor{gray41}{rgb}{0.41,0.41,0.41}
\definecolor{gray42}{rgb}{0.42,0.42,0.42}
\definecolor{gray43}{rgb}{0.43,0.43,0.43}
\definecolor{gray44}{rgb}{0.44,0.44,0.44}
\definecolor{gray45}{rgb}{0.45,0.45,0.45}
\definecolor{gray46}{rgb}{0.46,0.46,0.46}
\definecolor{gray47}{rgb}{0.47,0.47,0.47}
\definecolor{gray48}{rgb}{0.48,0.48,0.48}
\definecolor{gray49}{rgb}{0.49,0.49,0.49}
\definecolor{gray4}{rgb}{0.04,0.04,0.04}
\definecolor{gray50}{rgb}{0.50,0.50,0.50}
\definecolor{gray51}{rgb}{0.51,0.51,0.51}
\definecolor{gray52}{rgb}{0.52,0.52,0.52}
\definecolor{gray53}{rgb}{0.53,0.53,0.53}
\definecolor{gray54}{rgb}{0.54,0.54,0.54}
\definecolor{gray55}{rgb}{0.55,0.55,0.55}
\definecolor{gray56}{rgb}{0.56,0.56,0.56}
\definecolor{gray57}{rgb}{0.57,0.57,0.57}
\definecolor{gray58}{rgb}{0.58,0.58,0.58}
\definecolor{gray59}{rgb}{0.59,0.59,0.59}
\definecolor{gray5}{rgb}{0.05,0.05,0.05}
\definecolor{gray60}{rgb}{0.60,0.60,0.60}
\definecolor{gray61}{rgb}{0.61,0.61,0.61}
\definecolor{gray62}{rgb}{0.62,0.62,0.62}
\definecolor{gray63}{rgb}{0.63,0.63,0.63}
\definecolor{gray64}{rgb}{0.64,0.64,0.64}
\definecolor{gray65}{rgb}{0.65,0.65,0.65}
\definecolor{gray66}{rgb}{0.66,0.66,0.66}
\definecolor{gray67}{rgb}{0.67,0.67,0.67}
\definecolor{gray68}{rgb}{0.68,0.68,0.68}
\definecolor{gray69}{rgb}{0.69,0.69,0.69}
\definecolor{gray6}{rgb}{0.06,0.06,0.06}
\definecolor{gray70}{rgb}{0.70,0.70,0.70}
\definecolor{gray71}{rgb}{0.71,0.71,0.71}
\definecolor{gray72}{rgb}{0.72,0.72,0.72}
\definecolor{gray73}{rgb}{0.73,0.73,0.73}
\definecolor{gray74}{rgb}{0.74,0.74,0.74}
\definecolor{gray75}{rgb}{0.75,0.75,0.75}
\definecolor{gray76}{rgb}{0.76,0.76,0.76}
\definecolor{gray77}{rgb}{0.77,0.77,0.77}
\definecolor{gray78}{rgb}{0.78,0.78,0.78}
\definecolor{gray79}{rgb}{0.79,0.79,0.79}
\definecolor{gray7}{rgb}{0.07,0.07,0.07}
\definecolor{gray80}{rgb}{0.80,0.80,0.80}
\definecolor{gray81}{rgb}{0.81,0.81,0.81}
\definecolor{gray82}{rgb}{0.82,0.82,0.82}
\definecolor{gray83}{rgb}{0.83,0.83,0.83}
\definecolor{gray84}{rgb}{0.84,0.84,0.84}
\definecolor{gray85}{rgb}{0.85,0.85,0.85}
\definecolor{gray86}{rgb}{0.86,0.86,0.86}
\definecolor{gray87}{rgb}{0.87,0.87,0.87}
\definecolor{gray88}{rgb}{0.88,0.88,0.88}
\definecolor{gray89}{rgb}{0.89,0.89,0.89}
\definecolor{gray8}{rgb}{0.08,0.08,0.08}
\definecolor{gray90}{rgb}{0.90,0.90,0.90}
\definecolor{gray91}{rgb}{0.91,0.91,0.91}
\definecolor{gray92}{rgb}{0.92,0.92,0.92}
\definecolor{gray93}{rgb}{0.93,0.93,0.93}
\definecolor{gray94}{rgb}{0.94,0.94,0.94}
\definecolor{gray95}{rgb}{0.95,0.95,0.95}
\definecolor{gray96}{rgb}{0.96,0.96,0.96}
\definecolor{gray97}{rgb}{0.97,0.97,0.97}
\definecolor{gray98}{rgb}{0.98,0.98,0.98}
\definecolor{gray99}{rgb}{0.99,0.99,0.99}
\definecolor{gray9}{rgb}{0.09,0.09,0.09}
\definecolor{gray}{rgb}{0.75,0.75,0.75}
\definecolor{green1}{rgb}{0.00,1.00,0.00}
\definecolor{green2}{rgb}{0.00,0.93,0.00}
\definecolor{green3}{rgb}{0.00,0.80,0.00}
\definecolor{green4}{rgb}{0.00,0.55,0.00}
\definecolor{greenyellow}{rgb}{0.68,1.00,0.18}
\definecolor{green}{rgb}{0.00,1.00,0.00}
\definecolor{grey0}{rgb}{0.00,0.00,0.00}
\definecolor{grey100}{rgb}{1.00,1.00,1.00}
\definecolor{grey10}{rgb}{0.10,0.10,0.10}
\definecolor{grey11}{rgb}{0.11,0.11,0.11}
\definecolor{grey12}{rgb}{0.12,0.12,0.12}
\definecolor{grey13}{rgb}{0.13,0.13,0.13}
\definecolor{grey14}{rgb}{0.14,0.14,0.14}
\definecolor{grey15}{rgb}{0.15,0.15,0.15}
\definecolor{grey16}{rgb}{0.16,0.16,0.16}
\definecolor{grey17}{rgb}{0.17,0.17,0.17}
\definecolor{grey18}{rgb}{0.18,0.18,0.18}
\definecolor{grey19}{rgb}{0.19,0.19,0.19}
\definecolor{grey1}{rgb}{0.01,0.01,0.01}
\definecolor{grey20}{rgb}{0.20,0.20,0.20}
\definecolor{grey21}{rgb}{0.21,0.21,0.21}
\definecolor{grey22}{rgb}{0.22,0.22,0.22}
\definecolor{grey23}{rgb}{0.23,0.23,0.23}
\definecolor{grey24}{rgb}{0.24,0.24,0.24}
\definecolor{grey25}{rgb}{0.25,0.25,0.25}
\definecolor{grey26}{rgb}{0.26,0.26,0.26}
\definecolor{grey27}{rgb}{0.27,0.27,0.27}
\definecolor{grey28}{rgb}{0.28,0.28,0.28}
\definecolor{grey29}{rgb}{0.29,0.29,0.29}
\definecolor{grey2}{rgb}{0.02,0.02,0.02}
\definecolor{grey30}{rgb}{0.30,0.30,0.30}
\definecolor{grey31}{rgb}{0.31,0.31,0.31}
\definecolor{grey32}{rgb}{0.32,0.32,0.32}
\definecolor{grey33}{rgb}{0.33,0.33,0.33}
\definecolor{grey34}{rgb}{0.34,0.34,0.34}
\definecolor{grey35}{rgb}{0.35,0.35,0.35}
\definecolor{grey36}{rgb}{0.36,0.36,0.36}
\definecolor{grey37}{rgb}{0.37,0.37,0.37}
\definecolor{grey38}{rgb}{0.38,0.38,0.38}
\definecolor{grey39}{rgb}{0.39,0.39,0.39}
\definecolor{grey3}{rgb}{0.03,0.03,0.03}
\definecolor{grey40}{rgb}{0.40,0.40,0.40}
\definecolor{grey41}{rgb}{0.41,0.41,0.41}
\definecolor{grey42}{rgb}{0.42,0.42,0.42}
\definecolor{grey43}{rgb}{0.43,0.43,0.43}
\definecolor{grey44}{rgb}{0.44,0.44,0.44}
\definecolor{grey45}{rgb}{0.45,0.45,0.45}
\definecolor{grey46}{rgb}{0.46,0.46,0.46}
\definecolor{grey47}{rgb}{0.47,0.47,0.47}
\definecolor{grey48}{rgb}{0.48,0.48,0.48}
\definecolor{grey49}{rgb}{0.49,0.49,0.49}
\definecolor{grey4}{rgb}{0.04,0.04,0.04}
\definecolor{grey50}{rgb}{0.50,0.50,0.50}
\definecolor{grey51}{rgb}{0.51,0.51,0.51}
\definecolor{grey52}{rgb}{0.52,0.52,0.52}
\definecolor{grey53}{rgb}{0.53,0.53,0.53}
\definecolor{grey54}{rgb}{0.54,0.54,0.54}
\definecolor{grey55}{rgb}{0.55,0.55,0.55}
\definecolor{grey56}{rgb}{0.56,0.56,0.56}
\definecolor{grey57}{rgb}{0.57,0.57,0.57}
\definecolor{grey58}{rgb}{0.58,0.58,0.58}
\definecolor{grey59}{rgb}{0.59,0.59,0.59}
\definecolor{grey5}{rgb}{0.05,0.05,0.05}
\definecolor{grey60}{rgb}{0.60,0.60,0.60}
\definecolor{grey61}{rgb}{0.61,0.61,0.61}
\definecolor{grey62}{rgb}{0.62,0.62,0.62}
\definecolor{grey63}{rgb}{0.63,0.63,0.63}
\definecolor{grey64}{rgb}{0.64,0.64,0.64}
\definecolor{grey65}{rgb}{0.65,0.65,0.65}
\definecolor{grey66}{rgb}{0.66,0.66,0.66}
\definecolor{grey67}{rgb}{0.67,0.67,0.67}
\definecolor{grey68}{rgb}{0.68,0.68,0.68}
\definecolor{grey69}{rgb}{0.69,0.69,0.69}
\definecolor{grey6}{rgb}{0.06,0.06,0.06}
\definecolor{grey70}{rgb}{0.70,0.70,0.70}
\definecolor{grey71}{rgb}{0.71,0.71,0.71}
\definecolor{grey72}{rgb}{0.72,0.72,0.72}
\definecolor{grey73}{rgb}{0.73,0.73,0.73}
\definecolor{grey74}{rgb}{0.74,0.74,0.74}
\definecolor{grey75}{rgb}{0.75,0.75,0.75}
\definecolor{grey76}{rgb}{0.76,0.76,0.76}
\definecolor{grey77}{rgb}{0.77,0.77,0.77}
\definecolor{grey78}{rgb}{0.78,0.78,0.78}
\definecolor{grey79}{rgb}{0.79,0.79,0.79}
\definecolor{grey7}{rgb}{0.07,0.07,0.07}
\definecolor{grey80}{rgb}{0.80,0.80,0.80}
\definecolor{grey81}{rgb}{0.81,0.81,0.81}
\definecolor{grey82}{rgb}{0.82,0.82,0.82}
\definecolor{grey83}{rgb}{0.83,0.83,0.83}
\definecolor{grey84}{rgb}{0.84,0.84,0.84}
\definecolor{grey85}{rgb}{0.85,0.85,0.85}
\definecolor{grey86}{rgb}{0.86,0.86,0.86}
\definecolor{grey87}{rgb}{0.87,0.87,0.87}
\definecolor{grey88}{rgb}{0.88,0.88,0.88}
\definecolor{grey89}{rgb}{0.89,0.89,0.89}
\definecolor{grey8}{rgb}{0.08,0.08,0.08}
\definecolor{grey90}{rgb}{0.90,0.90,0.90}
\definecolor{grey91}{rgb}{0.91,0.91,0.91}
\definecolor{grey92}{rgb}{0.92,0.92,0.92}
\definecolor{grey93}{rgb}{0.93,0.93,0.93}
\definecolor{grey94}{rgb}{0.94,0.94,0.94}
\definecolor{grey95}{rgb}{0.95,0.95,0.95}
\definecolor{grey96}{rgb}{0.96,0.96,0.96}
\definecolor{grey97}{rgb}{0.97,0.97,0.97}
\definecolor{grey98}{rgb}{0.98,0.98,0.98}
\definecolor{grey99}{rgb}{0.99,0.99,0.99}
\definecolor{grey9}{rgb}{0.09,0.09,0.09}
\definecolor{grey}{rgb}{0.75,0.75,0.75}
\definecolor{honeydew1}{rgb}{0.94,1.00,0.94}
\definecolor{honeydew2}{rgb}{0.88,0.93,0.88}
\definecolor{honeydew3}{rgb}{0.76,0.80,0.76}
\definecolor{honeydew4}{rgb}{0.51,0.55,0.51}
\definecolor{honeydew}{rgb}{0.94,1.00,0.94}
\definecolor{hotpink}{rgb}{1.00,0.41,0.71}
\definecolor{indianred}{rgb}{0.80,0.36,0.36}
\definecolor{ivory1}{rgb}{1.00,1.00,0.94}
\definecolor{ivory2}{rgb}{0.93,0.93,0.88}
\definecolor{ivory3}{rgb}{0.80,0.80,0.76}
\definecolor{ivory4}{rgb}{0.55,0.55,0.51}
\definecolor{ivory}{rgb}{1.00,1.00,0.94}
\definecolor{khaki1}{rgb}{1.00,0.96,0.56}
\definecolor{khaki2}{rgb}{0.93,0.90,0.52}
\definecolor{khaki3}{rgb}{0.80,0.78,0.45}
\definecolor{khaki4}{rgb}{0.55,0.53,0.31}
\definecolor{khaki}{rgb}{0.94,0.90,0.55}
\definecolor{lavenderblush}{rgb}{1.00,0.94,0.96}
\definecolor{lavender}{rgb}{0.90,0.90,0.98}
\definecolor{lawngreen}{rgb}{0.49,0.99,0.00}
\definecolor{lemonchiffon}{rgb}{1.00,0.98,0.80}
\definecolor{lightblue}{rgb}{0.68,0.85,0.90}
\definecolor{lightcoral}{rgb}{0.94,0.50,0.50}
\definecolor{lightcyan}{rgb}{0.88,1.00,1.00}
\definecolor{lightgoldenrod}{rgb}{0.93,0.87,0.51}
\definecolor{lightgoldenrod}{rgb}{0.98,0.98,0.82}
\definecolor{lightgray}{rgb}{0.83,0.83,0.83}
\definecolor{lightgreen}{rgb}{0.56,0.93,0.56}
\definecolor{lightgrey}{rgb}{0.83,0.83,0.83}
\definecolor{lightpink}{rgb}{1.00,0.71,0.76}
\definecolor{lightsalmon}{rgb}{1.00,0.63,0.48}
\definecolor{lightsea}{rgb}{0.13,0.70,0.67}
\definecolor{lightsky}{rgb}{0.53,0.81,0.98}
\definecolor{lightslate}{rgb}{0.47,0.53,0.60}
\definecolor{lightslate}{rgb}{0.47,0.53,0.60}
\definecolor{lightslate}{rgb}{0.52,0.44,1.00}
\definecolor{lightsteel}{rgb}{0.69,0.77,0.87}
\definecolor{lightyellow}{rgb}{1.00,1.00,0.88}
\definecolor{limegreen}{rgb}{0.20,0.80,0.20}
\definecolor{linen}{rgb}{0.98,0.94,0.90}
\definecolor{magenta1}{rgb}{1.00,0.00,1.00}
\definecolor{magenta2}{rgb}{0.93,0.00,0.93}
\definecolor{magenta3}{rgb}{0.80,0.00,0.80}
\definecolor{magenta4}{rgb}{0.55,0.00,0.55}
\definecolor{magenta}{rgb}{1.00,0.00,1.00}
\definecolor{maroon1}{rgb}{1.00,0.20,0.70}
\definecolor{maroon2}{rgb}{0.93,0.19,0.65}
\definecolor{maroon3}{rgb}{0.80,0.16,0.56}
\definecolor{maroon4}{rgb}{0.55,0.11,0.38}
\definecolor{maroon}{rgb}{0.69,0.19,0.38}
\definecolor{mediumaquamarine}{rgb}{0.40,0.80,0.67}
\definecolor{mediumblue}{rgb}{0.00,0.00,0.80}
\definecolor{mediumorchid}{rgb}{0.73,0.33,0.83}
\definecolor{mediumpurple}{rgb}{0.58,0.44,0.86}
\definecolor{mediumsea}{rgb}{0.24,0.70,0.44}
\definecolor{mediumslate}{rgb}{0.48,0.41,0.93}
\definecolor{mediumspring}{rgb}{0.00,0.98,0.60}
\definecolor{mediumturquoise}{rgb}{0.28,0.82,0.80}
\definecolor{mediumviolet}{rgb}{0.78,0.08,0.52}
\definecolor{midnightblue}{rgb}{0.10,0.10,0.44}
\definecolor{mintcream}{rgb}{0.96,1.00,0.98}
\definecolor{mistyrose}{rgb}{1.00,0.89,0.88}
\definecolor{moccasin}{rgb}{1.00,0.89,0.71}
\definecolor{navajowhite}{rgb}{1.00,0.87,0.68}
\definecolor{navyblue}{rgb}{0.00,0.00,0.50}
\definecolor{navy}{rgb}{0.00,0.00,0.50}
\definecolor{oldlace}{rgb}{0.99,0.96,0.90}
\definecolor{olivedrab}{rgb}{0.42,0.56,0.14}
\definecolor{orange1}{rgb}{1.00,0.65,0.00}
\definecolor{orange2}{rgb}{0.93,0.60,0.00}
\definecolor{orange3}{rgb}{0.80,0.52,0.00}
\definecolor{orange4}{rgb}{0.55,0.35,0.00}
\definecolor{orangered}{rgb}{1.00,0.27,0.00}
\definecolor{orange}{rgb}{1.00,0.65,0.00}
\definecolor{orchid1}{rgb}{1.00,0.51,0.98}
\definecolor{orchid2}{rgb}{0.93,0.48,0.91}
\definecolor{orchid3}{rgb}{0.80,0.41,0.79}
\definecolor{orchid4}{rgb}{0.55,0.28,0.54}
\definecolor{orchid}{rgb}{0.85,0.44,0.84}
\definecolor{palegoldenrod}{rgb}{0.93,0.91,0.67}
\definecolor{palegreen}{rgb}{0.60,0.98,0.60}
\definecolor{paleturquoise}{rgb}{0.69,0.93,0.93}
\definecolor{paleviolet}{rgb}{0.86,0.44,0.58}
\definecolor{papayawhip}{rgb}{1.00,0.94,0.84}
\definecolor{peachpuff}{rgb}{1.00,0.85,0.73}
\definecolor{peru}{rgb}{0.80,0.52,0.25}
\definecolor{pink1}{rgb}{1.00,0.71,0.77}
\definecolor{pink2}{rgb}{0.93,0.66,0.72}
\definecolor{pink3}{rgb}{0.80,0.57,0.62}
\definecolor{pink4}{rgb}{0.55,0.39,0.42}
\definecolor{pink}{rgb}{1.00,0.75,0.80}
\definecolor{plum1}{rgb}{1.00,0.73,1.00}
\definecolor{plum2}{rgb}{0.93,0.68,0.93}
\definecolor{plum3}{rgb}{0.80,0.59,0.80}
\definecolor{plum4}{rgb}{0.55,0.40,0.55}
\definecolor{plum}{rgb}{0.87,0.63,0.87}
\definecolor{powderblue}{rgb}{0.69,0.88,0.90}
\definecolor{purple1}{rgb}{0.61,0.19,1.00}
\definecolor{purple2}{rgb}{0.57,0.17,0.93}
\definecolor{purple3}{rgb}{0.49,0.15,0.80}
\definecolor{purple4}{rgb}{0.33,0.10,0.55}
\definecolor{purple}{rgb}{0.63,0.13,0.94}
\definecolor{red1}{rgb}{1.00,0.00,0.00}
\definecolor{red2}{rgb}{0.93,0.00,0.00}
\definecolor{red3}{rgb}{0.80,0.00,0.00}
\definecolor{red4}{rgb}{0.55,0.00,0.00}
\definecolor{red}{rgb}{1.00,0.00,0.00}
\definecolor{rosybrown}{rgb}{0.74,0.56,0.56}
\definecolor{royalblue}{rgb}{0.25,0.41,0.88}
\definecolor{saddlebrown}{rgb}{0.55,0.27,0.07}
\definecolor{salmon1}{rgb}{1.00,0.55,0.41}
\definecolor{salmon2}{rgb}{0.93,0.51,0.38}
\definecolor{salmon3}{rgb}{0.80,0.44,0.33}
\definecolor{salmon4}{rgb}{0.55,0.30,0.22}
\definecolor{salmon}{rgb}{0.98,0.50,0.45}
\definecolor{sandybrown}{rgb}{0.96,0.64,0.38}
\definecolor{seagreen}{rgb}{0.18,0.55,0.34}
\definecolor{seashell1}{rgb}{1.00,0.96,0.93}
\definecolor{seashell2}{rgb}{0.93,0.90,0.87}
\definecolor{seashell3}{rgb}{0.80,0.77,0.75}
\definecolor{seashell4}{rgb}{0.55,0.53,0.51}
\definecolor{seashell}{rgb}{1.00,0.96,0.93}
\definecolor{sienna1}{rgb}{1.00,0.51,0.28}
\definecolor{sienna2}{rgb}{0.93,0.47,0.26}
\definecolor{sienna3}{rgb}{0.80,0.41,0.22}
\definecolor{sienna4}{rgb}{0.55,0.28,0.15}
\definecolor{sienna}{rgb}{0.63,0.32,0.18}
\definecolor{skyblue}{rgb}{0.53,0.81,0.92}
\definecolor{slateblue}{rgb}{0.42,0.35,0.80}
\definecolor{slategray}{rgb}{0.44,0.50,0.56}
\definecolor{slategrey}{rgb}{0.44,0.50,0.56}
\definecolor{snow1}{rgb}{1.00,0.98,0.98}
\definecolor{snow2}{rgb}{0.93,0.91,0.91}
\definecolor{snow3}{rgb}{0.80,0.79,0.79}
\definecolor{snow4}{rgb}{0.55,0.54,0.54}
\definecolor{snow}{rgb}{1.00,0.98,0.98}
\definecolor{springgreen}{rgb}{0.00,1.00,0.50}
\definecolor{steelblue}{rgb}{0.27,0.51,0.71}
\definecolor{tan1}{rgb}{1.00,0.65,0.31}
\definecolor{tan2}{rgb}{0.93,0.60,0.29}
\definecolor{tan3}{rgb}{0.80,0.52,0.25}
\definecolor{tan4}{rgb}{0.55,0.35,0.17}
\definecolor{tan}{rgb}{0.82,0.71,0.55}
\definecolor{thistle1}{rgb}{1.00,0.88,1.00}
\definecolor{thistle2}{rgb}{0.93,0.82,0.93}
\definecolor{thistle3}{rgb}{0.80,0.71,0.80}
\definecolor{thistle4}{rgb}{0.55,0.48,0.55}
\definecolor{thistle}{rgb}{0.85,0.75,0.85}
\definecolor{tomato1}{rgb}{1.00,0.39,0.28}
\definecolor{tomato2}{rgb}{0.93,0.36,0.26}
\definecolor{tomato3}{rgb}{0.80,0.31,0.22}
\definecolor{tomato4}{rgb}{0.55,0.21,0.15}
\definecolor{tomato}{rgb}{1.00,0.39,0.28}
\definecolor{turquoise1}{rgb}{0.00,0.96,1.00}
\definecolor{turquoise2}{rgb}{0.00,0.90,0.93}
\definecolor{turquoise3}{rgb}{0.00,0.77,0.80}
\definecolor{turquoise4}{rgb}{0.00,0.53,0.55}
\definecolor{turquoise}{rgb}{0.25,0.88,0.82}
\definecolor{violetred}{rgb}{0.82,0.13,0.56}
\definecolor{violet}{rgb}{0.93,0.51,0.93}
\definecolor{wheat1}{rgb}{1.00,0.91,0.73}
\definecolor{wheat2}{rgb}{0.93,0.85,0.68}
\definecolor{wheat3}{rgb}{0.80,0.73,0.59}
\definecolor{wheat4}{rgb}{0.55,0.49,0.40}
\definecolor{wheat}{rgb}{0.96,0.87,0.70}
\definecolor{whitesmoke}{rgb}{0.96,0.96,0.96}
\definecolor{white}{rgb}{1.00,1.00,1.00}
\definecolor{yellow1}{rgb}{1.00,1.00,0.00}
\definecolor{yellow2}{rgb}{0.93,0.93,0.00}
\definecolor{yellow3}{rgb}{0.80,0.80,0.00}
\definecolor{yellow4}{rgb}{0.55,0.55,0.00}
\definecolor{yellowgreen}{rgb}{0.60,0.80,0.20}
\definecolor{yellow}{rgb}{1.00,1.00,0.00}

\newcommand{\V}[1]{{\bf #1}}
\newcommand{\M}[1]{{\bf #1}}

\newcommand{\latentV}{\V{x}}
\newcommand{\measV}{\V{b}}
\newcommand{\sensM}{\M{A}}
\newcommand{\filterM}{\M{F}}
\newcommand{\filterV}{\V{f}}
\newcommand{\penaltyFunc}{\phi}
\newcommand{\weightNLS}{\tau}
\newcommand{\weightFidelity}{\lambda}
\newcommand{\weightSplit}{\rho}
\newcommand{\slackPrimaryV}{\V{z}}

\newcommand{\prox}{\V{prox}}
\newcommand{\derivativeOperator} {\nabla}

\newcommand{\diffusionFunc}{\psi}

\newcommand{\NLSprior}{\mathcal{S}}
\newcommand{\weightSplitNLS}{\rho_{s}}
\newcommand{\slackNLSV}{\V{v}}

\newcommand{\Fourier}{\mathcal{F}}
\newcommand{\invFourier}{\mathcal{F}^{-1}}
\newcommand{\transpose}{^\mathsf{T}}

\usepackage[pagebackref=true,breaklinks=true,letterpaper=true,colorlinks,bookmarks=false]{hyperref}

\iccvfinalcopy 

\ificcvfinal\fi
\begin{document}

\title{Discriminative Transfer Learning for General Image Restoration}

\author{Lei Xiao\\
University of British Columbia\\
\and
Felix Heide\\
Stanford University\\
\and
Wolfgang Heidrich\\
KAUST\\
\and
Bernhard Sch\"olkopf\\ 
MPI for Intelligent Systems\\
\and
Michael Hirsch\\
MPI for Intelligent Systems\\
}

\maketitle

\begin{abstract}
Recently, several discriminative learning approaches have been
proposed for effective image restoration, achieving convincing
trade-off between image quality and computational efficiency. However,
these methods require separate training for each restoration task
(e.g., denoising, deblurring, demosaicing) and problem condition
(e.g., noise level of input images). This makes it time-consuming and
difficult to encompass all tasks and conditions during training. In
this paper, we propose a discriminative transfer learning method
that incorporates formal proximal optimization and discriminative
learning for general image restoration. The method requires a
single-pass training and allows for reuse across various problems and
conditions while achieving an efficiency comparable to previous
discriminative approaches. Furthermore, after being trained, our model
can be easily transferred to new likelihood terms to solve untrained
tasks, or be combined with existing priors to further improve image
restoration quality.
\end{abstract}

\section{Introduction}
\label{sec:introduction}

Low-level vision problems, such as denoising, deconvolution and
demosaicing, have to be addressed as part of most imaging and vision
systems. Although a large body of work covers these classical
problems, low-level vision is still a very active area. The reason is
that, from a Bayesian perspective, solving them as statistical
estimation problems does not only rely on models for the likelihood
(i.e. the reconstruction task), but also on natural image priors as a
key component.

A variety of models for natural image statistics have been explored in
the past.  Traditionally, models for gradient
statistics~\cite{rudin1992nonlinear,krishnan2009fast}, including
total-variation, have been a popular choice. Another line of
works explores patch-based image statistics, either as per-patch
sparse model~\cite{elad2006image,zoran2011epll} or modeling non-local
similarity between patches~\cite{dabov2007image,dong2013nonlocally,gu2014weighted}.
These prior models are general in the sense that they can be applied
for various likelihoods, with the image formation and noise setting as
parameters. However, the resulting optimization problems are
prohibitively expensive, rendering them impractical for many real-time
tasks especially on mobile platforms.

Recently, a number of works~\cite{schmidt2014shrinkage,Chen_2015_CVPR}
have addressed this issue by truncating the iterative optimization and
learning discriminative image priors, tailored to a specific
reconstruction task (likelihood) and optimization approach. While
these methods allow to trade-off quality with the computational budget
for a given application, the learned models are highly specialized to
the image formation model and noise parameters, in contrast to
optimization-based approaches. Since each individual problem
instantiation requires costly learning and storing of the model
coefficients, current proposals for learned models are impractical for
vision applications with dynamically changing (often continuous)
parameters. This is a common scenario in most real-world vision
settings, as well as applications in engineering and scientific
imaging that rely on the ability to rapidly prototype methods.

In this paper, we combine discriminative learning techniques with
formal proximal optimization methods to learn generic models that can
be truly transferred across problem domains while achieving comparable
efficiency as previous discriminative approaches.  Using proximal
optimization
methods~\cite{geman1995nonlinear,parikh2013proximal,boyd2011distributed}
allows us to decouple the likelihood and prior which is key to learn
such shared models.  It also means that we can rely on well-researched
physically-motivated models for the likelihood, while learning priors
from example data. We verify our technique using the same model for a
variety of diverse low-level image reconstruction tasks and problem
conditions, demonstrating the effectiveness and versatility of our
approach. After training, our approach benefits from the proximal
splitting techniques, and can be naturally transferred to new
likelihood terms for untrained restoration tasks, or it can be
combined with existing state-of-the-art priors to further improve the
reconstruction quality. This is impossible with previous
discriminative methods. In particular, we make the following
contributions:

\begin{itemize}
\setlength\itemsep{0.1em}
\item We propose a discriminative transfer learning technique for
  general image restoration. It requires a single-pass training and
  transfers across different restoration tasks and problem conditions.
\item We show that our approach is general by demonstrating its
  robustness for diverse low-level problems, such as denoising,
  deconvolution, inpainting, and for varying noise settings.
\item We show that, while being general, our method achieves
  comparable computational efficiency as previous discriminative
  approaches, making it suitable for processing high-resolution images
  on mobile imaging systems.
\item We show that our method can naturally be combined with existing
  likelihood terms and priors after being trained. This allows our
  method to process untrained restoration tasks and take advantage of
  previous successful work on image priors (e.g., color and non-local
  similarity priors).
\end{itemize}


\section{Related work}
\label{sec:relatedwork}

Image restoration aims at computationally enhancing the quality of
images by undoing the adverse effects of image degradation such as
noise and blur. As a key area of image and signal processing it is an
extremely well studied problem and a plethora of methods exists, see
for example~\cite{milanfar2013tour} for a recent survey. Through the
successful application of machine learning and data-driven approaches,
image restoration has seen revived interest and much progress in
recent years. Broadly speaking, recently proposed
methods can be grouped into three classes: \emph{classical} approaches
that make no explicit use of machine learning, \emph{generative}
approaches that aim at probabilistic models of undegraded natural
images and \emph{discriminative} approaches that try to learn a direct
mapping from degraded to clean images. Unlike classical methods,
methods belonging to the latter two classes depend on the availability
of training data.

\smallskip
{\bf{Classical models}} focus on local image statistics and
aim at maintaining edges. Examples include total
variation~\cite{rudin1992nonlinear}, bilateral
filtering~\cite{tomasi2002bilateral} and anisotropic diffusion
models~\cite{weickert1998anisotropic}. More recent methods exploit the
non-local statistics of
images~\cite{buades2006review,dabov2007image,mairal2009non,dong2013nonlocally,gu2014weighted,talebi2014global}. In
particular the highly successful BM3D method~\cite{dabov2007image}
searches for similar patches within the same image and combines them
through a collaborative filtering step.

\smallskip
{\bf{Generative learning models}} seek to learn probabilistic
models of undegraded natural images. A simple, yet powerful subclass
include models that approximate the sparse gradient distribution of
natural
images~\cite{levin2007image,krishnan2009fast,krishnan2011}. More
expressive generative models include the fields of experts (FoE)
model~\cite{roth2009fields}, KSVD~\cite{elad2006image} and the EPLL
model~\cite{zoran2011epll}. While both FoE and KVSD learn a set of
filters whose responses are assumed to be sparse, EPLL models natural
images through Gaussian Mixture Models. All of these models have in
common that they are agnostic to the image restoration task, i.e. they
are transferable to any image degradation and can be combined in a
modular fashion with any likelihood and additional priors at test
time.

\smallskip
{\bf{Discriminative learning models}}  have recently become
increasingly popular for image restoration due to their attractive
tradeoff between high image restoration quality and efficiency at test
time. Methods include trainable random field models such as cascaded
shrinkage fields
(CSF)~\cite{schmidt2014shrinkage},
regression tree fields (RTF)~\cite{jancsary2012rtf}, trainable
nonlinear reaction diffusion (TRD) models~\cite{Chen_2015_CVPR}, as
well as deep convolutional networks~\cite{jain2009natural} and other
multi-layer perceptrons~\cite{burger2012image}.

Discriminative approaches owe their computational efficiency at
run-time to a particular feed-forward structure whose trainable
parameters are optimized for a particular task during training. Those
learned parameters are then kept fixed at test-time resulting in a
fixed computational cost. On the downside, discriminative models do
not generalize across tasks and typically necessitate separate
feed-forward architectures and separate training for each restoration
task (denoising, demosaicing, deblurring, etc.) as well as every
possible image degradation (noise level, Bayer pattern, blur kernel,
etc.).

\smallskip
In this work, we propose the \emph{discriminative transfer learning}
technique that is able to combine the strengths of both generative and
discriminative models: it maintains the flexibility of generative
models, but at the same time enjoys the computational efficiency of
discriminative models. While in spirit our approach is akin to the
recently proposed method of Rosenbaum and
Weiss~\cite{rosenbaum2015return}, who equipped the successful EPLL
model with a discriminative prediction step, the key idea in our
approach is to use proximal optimization
techniques~\cite{geman1995nonlinear,parikh2013proximal,boyd2011distributed}
that allow the decoupling of likelihood and prior and therewith share
the full advantages of a Bayesian generative modeling approach.

\begin{table}
\small
\centering
\caption{Analysis of state-of-the-art methods. In the table,
  ``Transferable'' means the model can be used for different
  restoration tasks and problem conditions; ``Modular'' means the
  method can be combined with other existing priors at test time.}
\label{tab:position}
\begin{tabular}{l|llllll}
               &FoE &EPLL &BM3D &TRD &ours\\\hline
Runtime efficiency &           &           &\checkmark &\checkmark &\checkmark\\
Easy to parallelize &      &           &           &\checkmark &\checkmark\\
Transferable &\checkmark &\checkmark &\checkmark &           &\checkmark\\
Modular  &\checkmark &\checkmark &\checkmark &           &\checkmark\\
\end{tabular}
\vspace{-0.1in}
\end{table}

Table~\ref{tab:position} summarizes the properties of the most
prominent state-of-the-art methods and puts our own proposed approach
into perspective.


\section{Proposed method}
\label{sec:method}

\subsection{Diversity of data likelihood}
\label{sec:diverse_likelihood}

\begin{figure*}
\centering
\includegraphics[width=1.9\columnwidth]{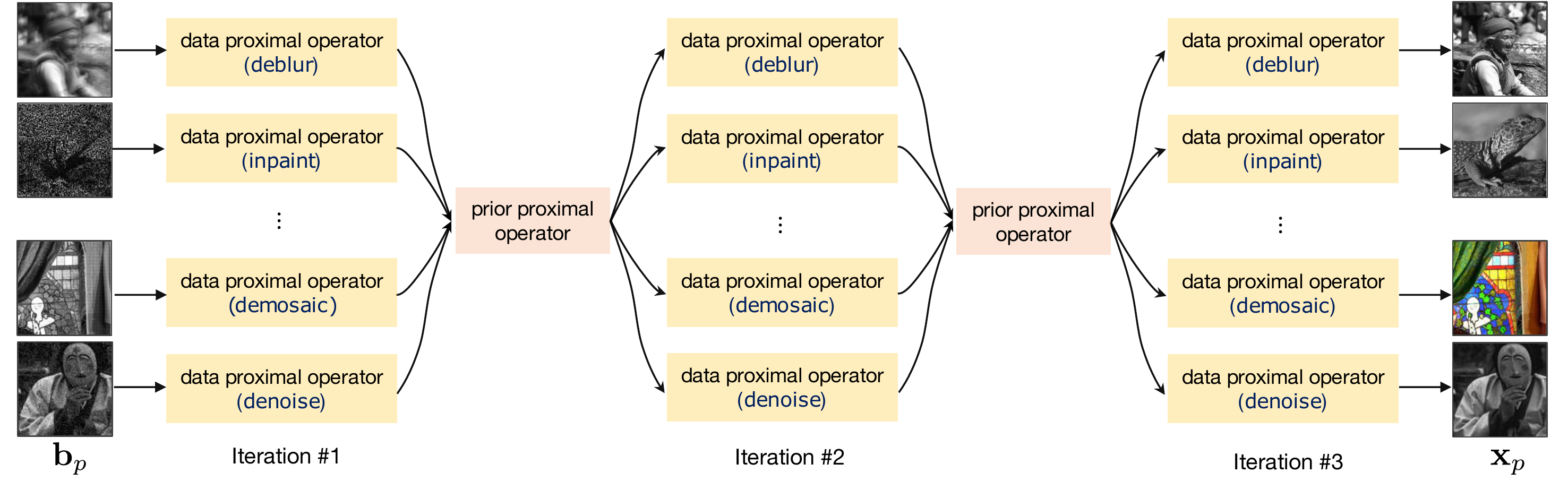}
\caption{The architecture of our method. Input images are drawn from various restoration tasks and problem conditions. Each iteration uses the same model parameters, forming a recurrent network.}
\label{fig:flowchart}
\end{figure*}

The seminal work of fields-of-experts (FoE)~\cite{roth2009fields}
generalizes the form of filter response based regularizers in the
objective function given in Eq.~\ref{eq:FoE_model}. The vectors
$\measV$ and $\latentV$ represent the observed and latent (desired)
image respectively, the matrix $\sensM$ is the sensing operator,
$\filterM_{i}$ represents 2D convolution with filter $\filterV_i$, and
$\penaltyFunc_i$ represents the penalty function on corresponding
filter responses $\filterM_{i}\latentV$. The positive scalar
$\weightFidelity$ controls the relative weight between the data
fidelity (likelihood) and the regularization term.
\begin{eqnarray}
\begin{aligned}
\label{eq:FoE_model}
\frac{\weightFidelity}{2} ||\measV - \sensM\latentV||^2_2 + \sum_{i=1}^{N} \penaltyFunc_{i}(\filterM_{i}\latentV)
\end{aligned}
\end{eqnarray}
\noindent The well-known anisotropic total-variation regularizer can be viewed as a special case of the FoE model where $\filterV_i$ is the derivative operator $\derivativeOperator$, and $\penaltyFunc_{i}$ the $\ell_1$ norm.

While there are various types of restoration tasks (e.g., denoising,
deblurring, demosaicing) and problem parameters (e.g., noise level of
input images), each problem has its own sensing matrix $\sensM$ and
optimal fidelity weight $\weightFidelity$. For example, $\sensM$ is an
identity matrix for denoising, a convolution operator for deblurring,
a binary diagonal matrix for demosaicing, and a random matrix for
compressive sensing~\cite{candes2008compressive}. $\weightFidelity$
depends on both the task and its parameters in order to produce the
best quality results.

The state-of-the-art discriminative learning methods
(CSF\cite{schmidt2014shrinkage}, TRD\cite{Chen_2015_CVPR}) derive an
end-to-end feed-forward model from Eq.~\ref{eq:FoE_model} for each
specific restoration task, and train this model to map the degraded
input images directly to the output. These methods have demonstrated
a great trade-off between high-quality and time-efficiency, however, as an inherent problem of the discriminative learning procedure, they require separate training for each restoration task and problem
condition. Given the diversity of data likelihood of image
restoration, this fundamental drawback of discriminative models makes it time-consuming and difficult to encompass
all tasks and conditions during training.

\subsection{Decoupling likelihood and prior}
\label{sec:decouple}

It is difficult to directly minimize Eq.~\ref{eq:FoE_model} when the penalty function $\penaltyFunc_i$ is non-linear and/or non-smooth (e.g., $\ell_p$ norm, $0<p\leq1$). Proximal algorithms~\cite{boyd2011distributed,geman1995nonlinear,chambolle2011first} instead relax Eq.~\ref{eq:FoE_model} and split the original problem into several easier subproblems that are solved alternately until convergence.

In this paper we employ the half-quadratic-splitting (HQS) algorithm~\cite{geman1995nonlinear} to relax Eq.~\ref{eq:FoE_model}, as it typically requires much fewer iterations to converge compared with other proximal methods such as ADMM~\cite{boyd2011distributed} and PD~\cite{chambolle2011first}. The relaxed objective function is given in Eq.~\ref{eq:FoE_split}:
\begin{eqnarray}
\begin{aligned}
\label{eq:FoE_split}
\frac{\weightFidelity}{2} ||\measV - \sensM\latentV||^2_2 + \frac{\weightSplit}{2} ||\slackPrimaryV - \latentV||^2_2 + \sum_{i=1}^{N} \penaltyFunc_i(\filterM_i \slackPrimaryV),
\end{aligned}
\end{eqnarray}
\noindent where a slack variable $\slackPrimaryV$ is introduced to approximate $\latentV$, and $\weightSplit$ is a positive scalar.

With the HQS algorithm, Eq.~\ref{eq:FoE_split} is iteratively
minimized by solving for the slack variable $\slackPrimaryV$ and the
latent image $\latentV$ alternately as in Eq.~\ref{eq:FoE_iter_z}
and~\ref{eq:FoE_iter_x} ($t=1,2,...,T$).
\begin{align}
&{\text{\bf Prior proximal operator:}}\nonumber\\
&\slackPrimaryV^t = \argmin_{\slackPrimaryV} \left( \frac{\weightSplit^t}{2} ||\slackPrimaryV - \latentV^{t-1}||^2_2 + \sum_{i=1}^{N} \penaltyFunc_i(\filterM_i \slackPrimaryV) \right) \label{eq:FoE_iter_z},\\
&{\text{\bf Data proximal operator:}}\nonumber\\
&\latentV^t = \argmin_{\latentV} \left( \weightFidelity ||\measV - \sensM\latentV||^2_2 + \weightSplit^t ||\slackPrimaryV^{t} - \latentV||^2_2 \right)\label{eq:FoE_iter_x},
\end{align}
\noindent where $\weightSplit^t$ increases as the iteration continues. This forces $\slackPrimaryV$ to become an increasingly good approximation of $\latentV$, thus making Eq.~\ref{eq:FoE_split} an increasingly good proxy for Eq.~\ref{eq:FoE_model}.

Note that, while most related approaches including
CSF~\cite{schmidt2014shrinkage} relax Eq.~\ref{eq:FoE_model} by
splitting on $\filterM_i \latentV$, we split on $\latentV$
instead. This is critical for deriving our approach. With this new
splitting strategy, the prior term and the data likelihood term in
the original objective Eq.~\ref{eq:FoE_model} are now separated into
two subproblems that we call the ``prior proximal operator''
(Eq.~\ref{eq:FoE_iter_z}) and the ``data
proximal operator'' (Eq.~\ref{eq:FoE_iter_x}), respectively.

\subsection{Discriminative transfer learning}
\label{sec:prox_field}

We observed that, while the data proximal operator in Eq.~\ref{eq:FoE_iter_x} is
task-dependent because both the sensing matrix $\sensM$ and fidelity
weight $\weightFidelity$ are problem-specific as explained in
Sec.~\ref{sec:diverse_likelihood}, the prior proximal-operator (i.e. $\slackPrimaryV^t$-update step in
Eq.~\ref{eq:FoE_iter_z}) is independent of the original restoration tasks and problem conditions.

This leads to our main insight: \emph{Discriminative learned models can be made transferable by using them in place of the prior proximal operator, embedded in a proximal optimization algorithm.}
This allows us to generalize a single discriminative learned model to a very
large class of problems, i.e. any linear inverse imaging problem,
while simultaneously overcoming the need for problem-specific
retraining. Moreover, it enables learning the task-dependent parameter $\weightFidelity$
in the data proximal operator for each problem in a single training pass, eliminating tedious hand-tuning at test time.

We also observed that, benefiting from our new splitting strategy, the prior proximal operator in Eq.~\ref{eq:FoE_iter_z} can be interpreted as a Gaussian denoiser on the intermediate image $\latentV^{t-1}$, since the least-squares consensus term is equivalent to a Gaussian
denoising term. This inspires us to utilize existing discriminative models that have been
successfully used for denoising (e.g. CSF, TRD).

For convenience, we denote the prior proximal operator as $\prox_{\Theta}$, i.e.
\begin{eqnarray}
\begin{aligned}
\label{eq:prox_fields}
\slackPrimaryV^t \coloneqq \prox_{\Theta} (\latentV^{t-1}, \weightSplit^t),
\end{aligned}
\end{eqnarray}
\noindent where the model parameter $\Theta$ includes a number of filters $\filterV_i$ and corresponding penalty functions $\penaltyFunc_i$. Inspired by the state-of-the-art discriminative
methods~\cite{schmidt2014shrinkage,Chen_2015_CVPR}, we propose to
learn the model $\prox_{\Theta}$, and the fidelity weight scalar
$\weightFidelity$, from training data. Recall that with our new
splitting strategy introduced in Sec.~\ref{sec:decouple}, the image
prior and data-fidelity term in the original objective
(Eq.~\ref{eq:FoE_model}) are contained in two separate subproblems
(Eq.~\ref{eq:FoE_iter_z} and~\ref{eq:FoE_iter_x}). This makes it
possible to train together an ensemble of diverse tasks (e.g.,
denoising, deblurring, or with different noise levels) each of which
has its own data proximal operator, while learning a single prior proximal operator $\prox_{\Theta}$ that is shared across tasks. This is in contrast to state-of-the-art discriminative methods such as CSF~\cite{schmidt2014shrinkage} and TRD~\cite{Chen_2015_CVPR} which train separate models for each task.

For clarity, in Fig.~\ref{fig:flowchart} we visualize the architecture
of our method. The input images may represent various restoration
tasks and problem conditions. At each HQS iteration, each image
$\latentV^{t}_p$ from problem $p$ is updated by {\emph{its own}} data
proximal operator in Eq.~\ref{eq:FoE_iter_x} which contains separate
trainable fidelity weight $\weightFidelity_p$ and pre-defined sensing
matrix $\sensM_p$; then each slack image $\slackPrimaryV^t_p$ is
updated by the same, {\emph{shared}} prior proximal operator implemented by
a learned, discriminative model.

\smallskip
{\noindent \bf Recurrent network.} Note that in
Fig.~\ref{fig:flowchart} each HQS iteration uses exactly the same
model parameters, forming a recurrent network. This is in contrast to
previous discriminative learning methods including CSF and TRD, which
form feed-forward networks. Our recurrent network architecture
maintains the convergence property of the proximal optimization
algorithm (HQS), and is critical for our method to transfer
between various tasks and problem conditions.

\smallskip
{\noindent \bf {Shared prior proximal operator.}}
While {\emph{any}} discriminative Gaussian denoising model could be
used as $\prox_\Theta$ in our framework, we
specifically propose to use the multi-stage non-linear diffusion
process that is modified from the TRD~\cite{Chen_2015_CVPR} model, for
its efficiency. The model is given in Eq.~\ref{eq:tnrd_denoise}.
\vspace{-0.2in}
\begin{eqnarray}
\begin{aligned}
\label{eq:tnrd_denoise}
&\slackPrimaryV^t_k = \slackPrimaryV^t_{k-1} - \sum_{i=1}^N{\filterM^k_{i}}\transpose\diffusionFunc^k_{i}(\filterM^k_{i} \slackPrimaryV^t_{k-1}),\\
&s.t. \quad \slackPrimaryV^t_0 = \latentV^{t-1}, \quad k=1,2,...,K,
\end{aligned}
\end{eqnarray}
where $k$ is the stage index, filters $\filterM^k_{i}$, function $\diffusionFunc^k_{i}$ are trainable model parameters at each stage, and $\slackPrimaryV^t_0$ is the initial value of $\slackPrimaryV^t_k$. Note that, different from TRD, our model does not contain the reaction term which would be $-\weightSplit^t \alpha_k(\slackPrimaryV^t_{k-1} - \latentV^{t-1})$ with step size $\alpha_k$. The main reasons for this modification are:
\begin{itemize}
\setlength\itemsep{0.1em}
   \item The data constraint is contained in $\latentV^t$ update in Eq.~\ref{eq:FoE_iter_x};
   \item More importantly, by dropping the reaction term our model gets rid of the weight $\weightSplit^t$ which changes at each HQS iteration. Therefore, our proximal operator $\prox_\Theta(\latentV^{t-1}, \weightSplit^t)$ is simplified to be:
\begin{eqnarray}
\begin{aligned}
\label{eq:prox_fields_final}
\slackPrimaryV^t \coloneqq \prox_{\Theta} (\latentV^{t-1})
\end{aligned}
\end{eqnarray}
\end{itemize}

The parameter $\Omega$ to learn in our method includes
$\weightFidelity$'s for each problem class $p$ (restoration task and
problem condition), and $\Theta=\{\filterM^k_{i}, \diffusionFunc^k_{i}\}$ in the prior proximal operator
shared across different classes, i.e. $\Omega=\{\weightFidelity_p,
\Theta\}$. Even though the scalar parameters $\weightFidelity_p$ are
trained, our method allows users to override them at test time to
handle non-trained problem classes or specific
inputs as we will show in Sec.~\ref{sec:results}. This contrasts to previous discriminative approaches whose
model parameters are all fixed at test time. The subscript $p$
indicating the problem class in $\weightFidelity_p$ is omitted below
for convenience. The values of $\weightSplit^t$ are pre-selected:
$\weightSplit^1=1$ and $\weightSplit^t=2\weightSplit^{t-1}$ for $t>1$.

\begin{algorithm}[ht]
\caption{Proposed algorithm}
\label{alg:alg_main}
\begin{algorithmic}[1]
\INPUT degraded image $\measV$
\OUTPUT recovered image $\latentV$
\STATE $\latentV^{0} = \measV, \weightSplit^1 = 1$ {\em{(initialization)}}\\
\FOR{$t=1$ to $T$}
\STATE {\em{(Update $\slackPrimaryV^{t}$ by Eq.~\ref{eq:tnrd_denoise} below)}}\\
\STATE $\slackPrimaryV^t_0 = \latentV^{t-1}$\\
\FOR{$k=1$ to $K$}
\STATE $\slackPrimaryV^t_k = \slackPrimaryV^t_{k-1} - \sum_{i=1}^N{\filterM^k_{i}}\transpose\diffusionFunc^k_{i}(\filterM^k_{i} \slackPrimaryV^t_{k-1})$\\
\ENDFOR
\STATE $\slackPrimaryV^t = \slackPrimaryV^t_K$\\
\STATE {\em{(Update $\latentV^{t}$ by Eq.~\ref{eq:FoE_iter_x} below)}}\\
\STATE $\latentV^t = \argmin_{\latentV} \weightFidelity ||\measV - \sensM\latentV||^2_2 + \weightSplit^t ||\slackPrimaryV^{t} - \latentV||^2_2
$\\
\STATE $\weightSplit^{t+1}=2\weightSplit^{t}$\\
\ENDFOR
\end{algorithmic}
\end{algorithm}

Note that a multi-stage model as in Eq.~\ref{eq:tnrd_denoise} is not possible if we split on $\filterM_i \latentV$ instead of $\latentV$ in Eq.~\ref{eq:FoE_model} and~\ref{eq:FoE_split}. For clarity, an overview of the proposed algorithm is given in Algorithm~\ref{alg:alg_main}.

\subsection{Training}
\label{sec:learning}

We consider denoising and deconvolution tasks at training, where the sensing operator $\sensM$ is an identity matrix, or a block circulant matrix with circulant blocks that represents 2D convolution with randomly drawn blur kernels respectively.
In denoising tasks, the $\latentV^t$ update in Eq.~\ref{eq:FoE_iter_x} has a closed-form solution:
\begin{eqnarray}
\begin{aligned}
\label{eq:FoE_iter_x_denoise}
\latentV^t = (\weightFidelity \measV + \rho^t\slackPrimaryV^t)/(\weightFidelity+\rho^t)
\end{aligned}
\end{eqnarray}
\noindent In deconvolution tasks, the $\latentV^t$ update in
Eq.~\ref{eq:FoE_iter_x} has a closed-form solution in the Fourier domain:

\begin{eqnarray}
\begin{aligned}
\label{eq:FoE_iter_x_deblur}
\latentV^t = \invFourier\left(\frac{\Fourier(\weightFidelity \sensM\transpose\measV + \weightSplit^t \slackPrimaryV^t)}{\Fourier(\weightFidelity \sensM\transpose\sensM + \weightSplit^t)}\right),
\end{aligned}
\end{eqnarray}
\noindent where $\Fourier$ and $\invFourier$ represent Fourier and
inverse Fourier transform respectively. Note that, compared to
CSF~\cite{schmidt2014shrinkage}, our method does not require FFT
computations for denoising tasks. We use the L-BFGS
solver~\cite{schmidt05minfunc} with analytic gradient computation for
training. The training loss function $\ell$ is defined as the negative
average Peak Signal-to-Noise Ratio (PSNR) of reconstructed images. The
gradient of $\ell$ w.r.t. the model parameters
$\Omega=\{\weightFidelity_p, \Theta\}$ is computed by accumulating
gradients at all HQS iterations, i.e.
\begin{eqnarray}
\begin{aligned}
\label{eq:bp_gradient_model_iter}
\frac{\partial \ell}{\partial \Omega} = \sum_{t=1}^{T} \left(\frac{\partial \latentV^t}{\partial \weightFidelity} \frac{\partial \ell}{\partial \latentV^t} + \frac{\partial \slackPrimaryV^t}{\partial \Theta} \frac{\partial \latentV^t}{\partial \slackPrimaryV^t}\frac{\partial \ell}{\partial \latentV^t}\right).
\end{aligned}
\end{eqnarray}
\noindent The 1D functions $\diffusionFunc^k_i$ in Eq.~\ref{eq:tnrd_denoise} are parameterized as a linear combination of equidistant-positioned Gaussian kernels whose weights are trainable.

\begin{figure}
\centering
\subfigure[Filters at stage 1.]
{\includegraphics[width=.7\columnwidth]{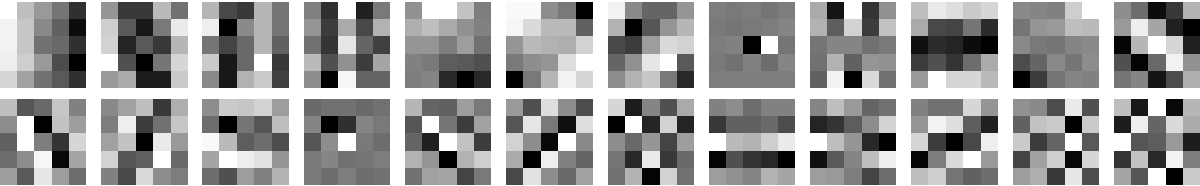}}\\
\vspace{-0.1in}
\subfigure[Filters at stage 2.]
{\includegraphics[width=.7\columnwidth]{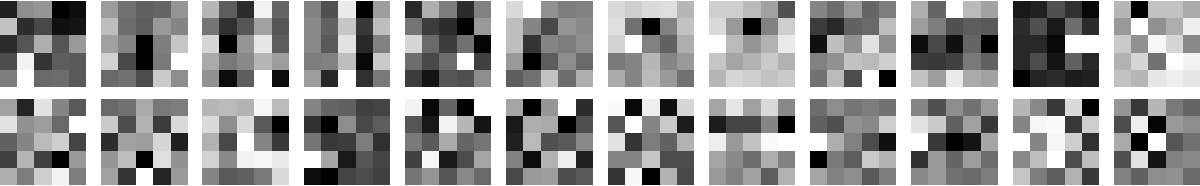}}\\
\vspace{-0.1in}
\subfigure[Filters at stage 3.]
{\includegraphics[width=.7\columnwidth]{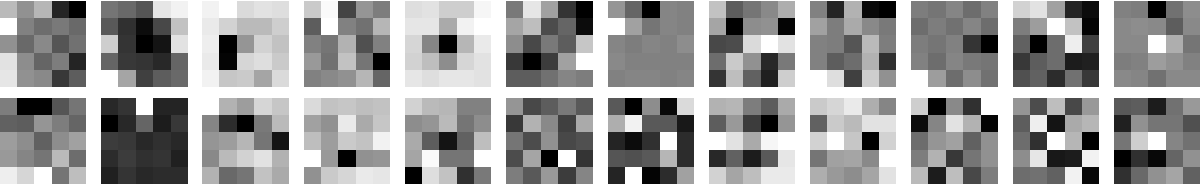}}\\
\caption{Trained filters at each stage ($k$ in Eq.~\ref{eq:tnrd_denoise}) of the proximal operator $\prox_\Theta$ in our model (3 stages each with 24 5$\times$5 filters).}
\label{fig:vis_filters}
\end{figure}

\smallskip
\noindent{\bf{Progressive training.}} A progressive scheme is proposed
to make the training more effective. First, we set the number of HQS
iterations to be 1, and train $\weightFidelity's$ and the model
$\Theta$ of each stage in $\prox_{\Theta}$ in a greedy fashion. Then,
we gradually increase the number of HQS iterations from 1 to $T$ where
at each step the model $\Omega=\{\weightFidelity, \Theta\}$ is refined
from the result of the previous step. The L-BFGS iterations are set to be
200 for the greedy training steps, and 100 for the refining
steps. Fig.~\ref{fig:vis_filters} shows examples of learned
filters in $\prox_{\Theta}$.

\section{Results}
\label{sec:results}

\begin{figure}
\vspace{-0.1in}
\centering
{\includegraphics[width=.75\columnwidth]{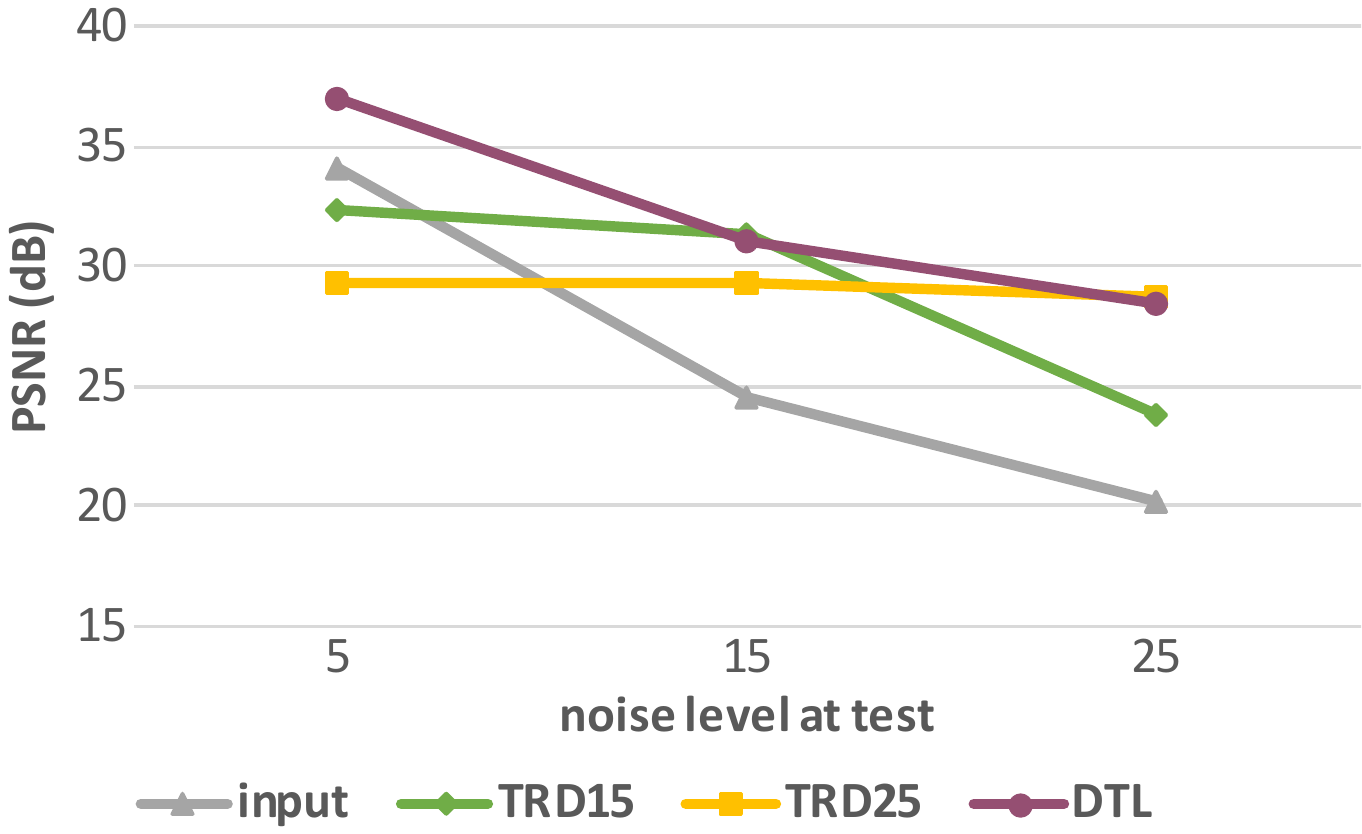}}\\
\caption{Analysis of model generality on image denoising. In this plot, ``TRD15'' denotes the TRD model trained at noise $\sigma=15$, and ``TRD25'' trained at noise $\sigma=25$. Our model DTL is trained with mixed noise levels in a single pass.}
\label{fig:analysis_generic_denoise}
\end{figure}

\noindent{\bf{Denoising and generality analysis.}} We compare the
proposed discriminative transfer learning (DTL) method
with state-of-the-art image denoising techniques,
including KSVD~\cite{elad2006image}, FoE~\cite{roth2009fields},
BM3D~\cite{dabov2007image}, LSSC~\cite{mairal2009non},
WNNM~\cite{gu2014weighted}, EPLL~\cite{zoran2011epll},
opt-MRF~\cite{chen2013revisiting}, ARF~\cite{barbu2009training},
CSF~\cite{schmidt2014shrinkage} and TRD~\cite{Chen_2015_CVPR}. The
subscript in CSF$_5$ and TRD$_5$ indicates the number of cascaded
stages (each stage has different model parameters). The subscript and
superscript in our method DTL$^5_3$ indicate the number of diffusion
stages ($K=3$ in Algorithm~\ref{alg:alg_main}) in our proximal
operator $\prox_{\Theta}$, and the number of HQS iterations ($T=5$ in
Alg.~\ref{alg:alg_main}), respectively. Note that the complexity
(size) of our model is linear in $K$, but independent of $T$. CSF, TRD
and DTL use 24 filters of size 5$\times$5 pixels at all stages in this
section.

The compared discriminative methods, CSF$_5$ and TRD$_5$ both are trained at single noise level $\sigma=15$ that is the same as the test images. In contrast, our model is trained on 400 images (100$\times$100 pixels) cropped from~\cite{roth2009fields} with random and discrete noise levels (standard deviation $\sigma$) varying between 5 and 25. The images with the same noise level share the same data fidelity weight $\weightFidelity$ at training.

To verify the generality of our method on varying noise levels, we test our model DTL$^3_3$ (trained with varying noise levels in a single pass) and two TRD models (trained at specific noise levels 15 and 25) on 3 sets of 68 images with noise $\sigma=5,15,25$ respectively. The average PSNR values are shown in Fig.~\ref{fig:analysis_generic_denoise}. Although performing slightly below the TRD model trained for the exact noise level used at test time, our method is more generic and works robustly for various noise levels. The performance of the discriminative TRD method drops down quickly as the problem condition (i.e. noise level) at test differs from its training data. In sharp contrast to discriminative methods (CSF, TRD, etc), which are inherently specialized for a given problem setting, i.e. noise level, the proposed approach transfers across different problem settings. More analysis can be found in the supplementary material.

All compared methods are evaluated on the 68 test images from~\cite{roth2009fields} and the averaged PSNR values are reported in Table~\ref{tab:psnr_denoise15}. The compared discriminative methods (CSF, TRD, etc) were trained for exactly the same noise level as the test images (i.e. the best case for them), while our model was trained with mixed noise levels and works robustly for arbitrary noise levels.
Our results are comparable to generic methods such as KSVD, FoE and BM3D, and very close to discriminative methods such as CSF$_5$, while at the same time being much more time-efficient.

\begin{table}
\centering
\caption{Average PSNR(dB) on 68 images from~\cite{roth2009fields} for denoising.}
\label{tab:psnr_denoise15}
\small
\vspace{0.02in}
\begin{tabular}{llllll}
\hline
\rowcolor[HTML]{EFEFEF}
KSVD    & FoE   & BM3D  & LSSC  & WNNM & EPLL  \\
30.87   & 30.99 & 31.08 & 31.27 & 31.37 & 31.19 \\ 
\hline
\rowcolor[HTML]{EFEFEF}
opt-MRF & ARF   & CSF$_5$   & TRD$_5$   & DTL$^3_3$ & DTL$^5_3$\\
31.18   & 30.70 & 31.14     & 31.30     & 30.91       & 31.00 \\ 
\hline 
\end{tabular}
\end{table}

\begin{table}
\caption{Runtime (seconds) comparison for image denoising.}
\label{tab:runtime}
\small
\resizebox{\linewidth}{!} {
\begin{tabular}{llllll}
\hline
Image size & $256^2$   & $512^2$   & $1024^2$   & $2048^2$ & $4096^2$ \\ \hline
WNNM       & 157.73  & 657.75   & 2759.79  & -      & -      \\
EPLL       & 29.21   & 111.52    &463.71   &-    &-     \\
BM3D       & 0.78    & 3.45  &15.24  &62.81    & 275.39      \\
CSF$_5$    & 1.23   & 2.22  & 7.35 & 27.08   &93.66      \\
TRD$_5$    &0.39  & 0.71 &  2.01 &7.57  & 29.09        \\
DTL$^3_3$ & 0.60 & 1.19  & 3.45 & 12.97  & 56.19   \\
DTL$^3_3$ (Halide) & 0.11  & 0.26 & 1.60 & 5.61 & 20.85 \\\hline
\end{tabular}}
\end{table}

\smallskip
\noindent{\bf{Run-time comparison.}} In Table~\ref{tab:runtime} we
compare the run-time of our method and state-of-the-art methods. The
experiments were performed on a laptop computer with Intel i7-4720HQ
CPU and 16GB RAM. WNNM and EPLL ran out-of-memory for images over 4
megapixels in our experiments. CSF$_5$, TRD$_5$ and DTL$^3_3$ all use
``parfor'' setting in Matlab. DTL$^3_3$ is significantly faster than
all compared generic methods (WNNM, EPLL, BM3D) and even the
discriminative method CSF$_5$. Run-time of DTL$^3_3$ is about 1.5
times that of TRD$_5$, which is expected as they use 5 versus 9
diffusion steps in total. In addition, we implement our method in Halide
language~\cite{ragan2013halide}, which has become popular recently for
high-performance image processing applications, and report the
run-time on the same CPU as mentioned above.

\smallskip
\noindent{\bf{Deconvolution.}} In this experiment, we train
a model with an ensemble of denoising and deconvolution tasks on 400
images (100$\times$100 pixels) cropped from~\cite{roth2009fields}, in
which 250 images are generated for denoising tasks with random noise
levels $\sigma$ varying between 5 and 25, and the other 150 images are
generated by blurring the images with random 25$\times$25 kernels
(PSFs) and then adding Gaussian noise with $\sigma$ ranging between 1 and
5. All images are quantized to 8 bits.

\begin{figure}
\centering
{\includegraphics[width=.9\columnwidth]{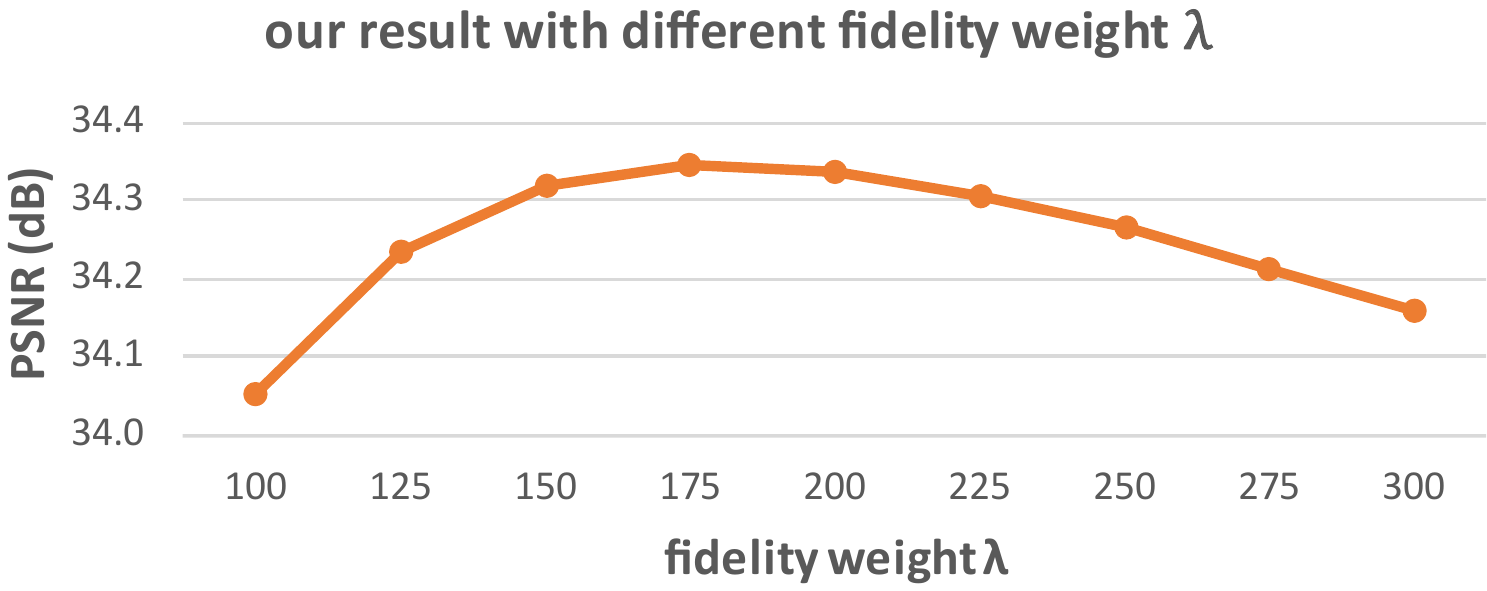}}\\
\caption{Our results with different fidelity weight $\weightFidelity$ for the non-blind deconvolution experiment reported in Table~\ref{tab:psnr_deblur}.}
\label{fig:plot_deblur_different_lambda}
\end{figure}

We compare our method with state-of-the-art non-blind deconvolution methods including Levin et al.~\cite{levin2007image}, Schmidt et al.~\cite{schmidt2013discriminative} and CSF~\cite{schmidt2014shrinkage}. Note that TRD~\cite{Chen_2015_CVPR} does not support non-blind deconvolution. We test the methods on the benchmark dataset from~\cite{levin2011efficient} which contains 32 real-captured images and report the average PSNR values in Table~\ref{tab:psnr_deblur}. The results of compared methods are quoted from~\cite{schmidt2014shrinkage}.

As said in Sec.~\ref{sec:prox_field}, while the scalar weight $\weightFidelity$ is trained, our method allows users to override it at test time for untrained problem classes or specific inputs. Fig.~\ref{fig:plot_deblur_different_lambda} shows our results with different $\weightFidelity$ on the experiments compared in Table~\ref{tab:psnr_deblur}. Within a fairly wide range of $\weightFidelity$, our method outperforms all previous methods.

We further test the above model trained with ensemble tasks on the denoising experiment in Table~\ref{tab:psnr_denoise15}. The result average PSNR is 30.98dB, which is comparable to the result with the model trained only on the denoising task.

\begin{table}
\centering
\caption{Average PSNR (dB) on 32 images from~\cite{levin2011efficient} for non-blind deconvolution.}
\vspace{0.02in}
\label{tab:psnr_deblur}
\begin{tabular}{llllll}
\hline
\rowcolor[HTML]{EFEFEF}
Input & Levin~\cite{levin2007image}    & Schmidt~\cite{schmidt2013discriminative}   & CSF$^{\text{pw}}_3$  & DTL$^3_3$\\
22.86 & 32.73   & 33.97 & 33.48 & 34.34 \\
\hline
\end{tabular}
\vspace{-0.1in}
\end{table}

\smallskip
\noindent{\bf{Modularity with existing priors.}} As shown above, even
though the fidelity weight $\weightFidelity$ is trainable, our method
allows users to override its value at test time.
This property also
makes it possible to combine our model
(after being trained) with existing state-of-the-art priors at test
time, in which case $\weightFidelity$ typically needs to be
adjusted. This allows our method to take advantage of previous successful
work on image priors. Again, this is not possible with previous discriminative
methods (CSF, TRD).

\begin{figure}
\centering
\subfigure[Input (20.17dB)]
{\includegraphics[width=.45\columnwidth]{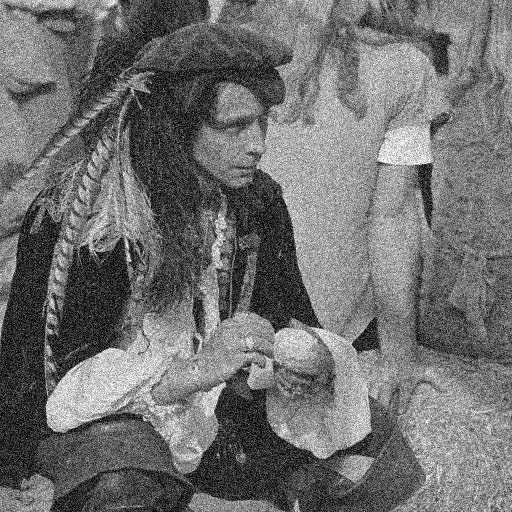}}
\subfigure[BM3D (29.62dB)]
{\includegraphics[width=.45\columnwidth]{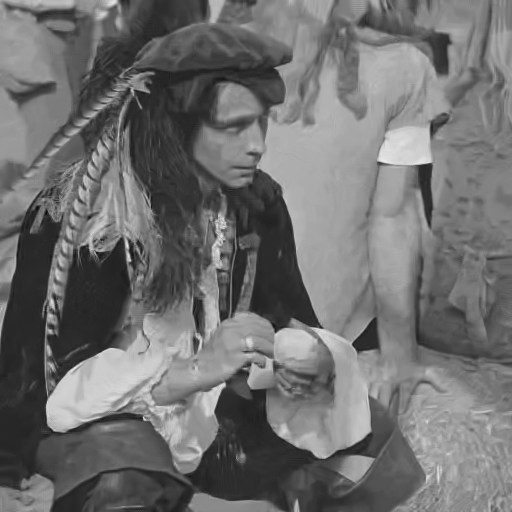}}\\
\vspace{-0.1in}
\subfigure[DTL$^5_3$ (29.48dB)]
{\includegraphics[width=.45\columnwidth]{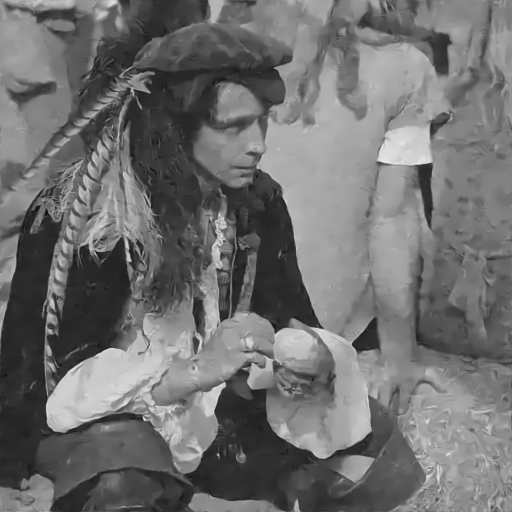}}
\subfigure[DTL$^5_3$ + BM3D (29.74dB)]
{\includegraphics[width=.45\columnwidth]{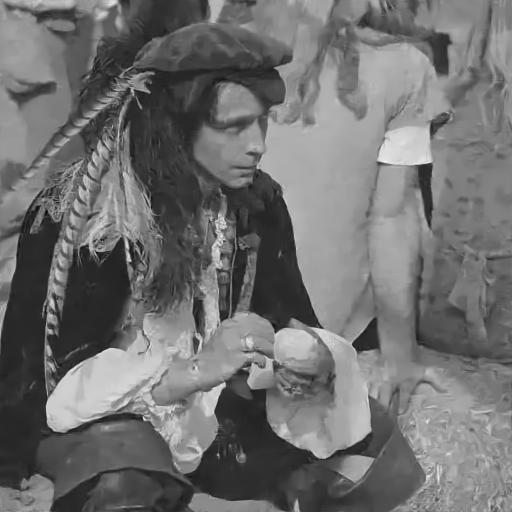}}
\caption{Experiment on incorporating non-local patch similarity prior (BM3D) with our model after being trained. The input noise level $\sigma=25$. Please zoom in for better view.}
\label{fig:result_collaborative}
\vspace{-0.1in}
\end{figure}

In Fig.~\ref{fig:result_collaborative} we show an example to incorporate a non-local patch similarity prior (BM3D~\cite{dabov2007image}) with our method to further improve the denoising quality. BM3D performs well in removing noise especially in smooth regions but usually over-smoothes edges and textures. Our original model (DTL$^5_3$) well preserves sharp edges however sometimes introduces artifacts in smooth regions when the input noise level is high. By combining those two methods, which is easy with our HQS framework, the result is improved both visually and quantitatively.

We give the derivation of the proposed hybrid method below. Let $\NLSprior(\latentV)$ represents the non-local patch similarity prior. The objective function is:
\begin{eqnarray}
\begin{aligned}
\label{eq:obj_combine_bm3d}
\frac{\weightFidelity}{2} ||\measV - \sensM\latentV||^2_2 + \sum_{i=1}^{N} \penaltyFunc_{i}(\filterM_{i}\latentV) + \weightNLS \NLSprior(\latentV)
\end{aligned}
\end{eqnarray}
\noindent Applying the HQS technique described in Sec.~\ref{sec:method}, we relax the objective to be:
\begin{eqnarray}
\begin{aligned}
\label{eq:obj_combine_bm3d_split}
\frac{\weightFidelity}{2} ||\measV - \sensM\latentV||^2_2 &+ \frac{\weightSplit}{2} ||\slackPrimaryV - \latentV||^2_2 + \sum_{i=1}^{N} \penaltyFunc_i(\filterM_i \slackPrimaryV)\\
&+ \frac{\weightSplitNLS}{2} ||\slackNLSV - \latentV||^2_2 + \weightNLS \NLSprior(\slackNLSV)
\end{aligned}
\end{eqnarray}
\noindent Then we minimize Eq.~\ref{eq:obj_combine_bm3d_split} by alternately solving the following 3 subproblems:
\begin{equation}
\begin{aligned}
&\slackPrimaryV^t = \prox_{\Theta} (\latentV^{t-1})\\
&\slackNLSV^t = \argmin_{\slackNLSV} \frac{\weightSplitNLS^t}{2} ||\slackNLSV - \latentV^{t-1}||^2_2 + \weightNLS \NLSprior(\slackNLSV)\approx {\text{BM3D}}(\latentV^{t-1}, \frac{\weightNLS}{\weightSplitNLS^t}) \\
&\latentV^t = \argmin_{\latentV} \weightFidelity ||\measV - \sensM\latentV||^2_2 + \weightSplit^t ||\slackPrimaryV^{t} - \latentV||^2_2 + \weightSplitNLS^t ||\slackNLSV^t - \latentV||^2_2,
\label{eq:combine_bm3d_iter}
\end{aligned}
\end{equation}
\noindent where $\prox_{\Theta}$ is from our previous training, and the $\slackNLSV^t$ subproblem is {\em{approximated}} by running BM3D software on $\latentV^{t-1}$ with noise parameter $\frac{\weightNLS}{\weightSplitNLS^t}$ following~\cite{Venkatakrishnan2013PlugandPlayPF,heide2014flexisp}.

\begin{figure}
\centering
\subfigure[Ground truth]
{\includegraphics[width=.48\columnwidth]{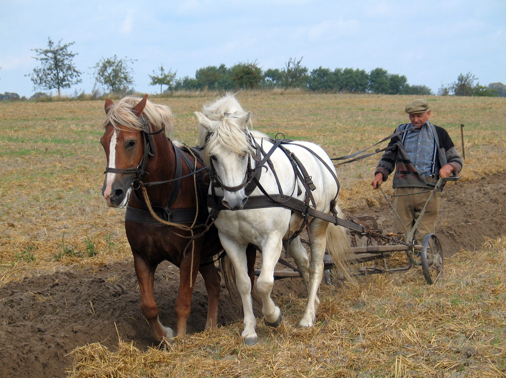}}
\subfigure[Input (20.18dB)]
{\includegraphics[width=.48\columnwidth]{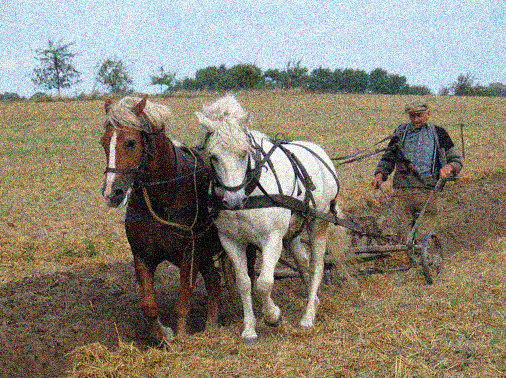}}\\
\vspace{-0.1in}
\subfigure[TRD$_5$ (28.06dB)]
{\includegraphics[width=.48\columnwidth]{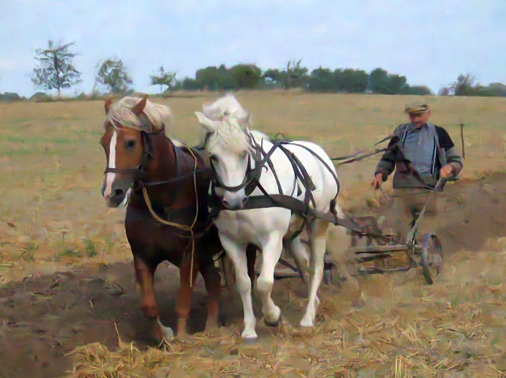}}
\subfigure[DTL$^5_3$ (27.80dB)]
{\includegraphics[width=.48\columnwidth]{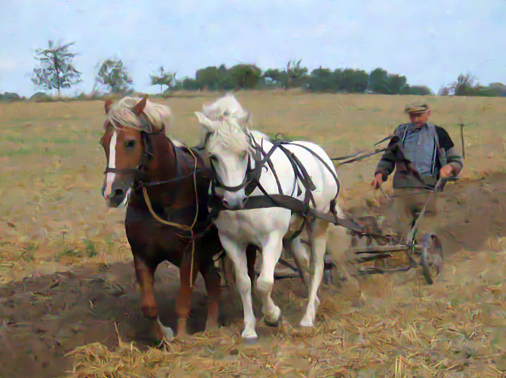}}\\
\vspace{-0.1in}
\subfigure[TV + cross (26.89dB)]
{\includegraphics[width=.48\columnwidth]{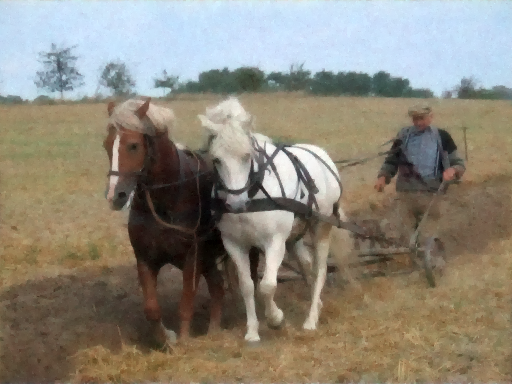}}
\subfigure[DTL$^5_3$ + cross (28.69dB)]
{\includegraphics[width=.48\columnwidth]{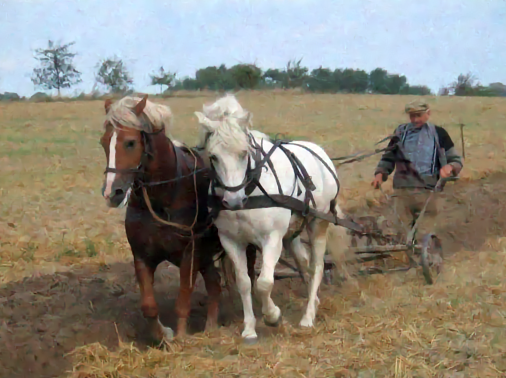}}
\caption{Experiment on incorporating a color prior~\cite{heide2014flexisp} with our model after being trained. The input noise level $\sigma=25$. (e,f) show the results by combining total variation (TV) denoising with a cross-channel prior, and our method with cross-channel prior, respectively. Please zoom in for better view.}
\label{fig:result_color}
\vspace{-0.1in}
\end{figure}

Similarly, our method can incorporate color image priors (e.g., cross-channel edge-concurrence prior~\cite{heide2014flexisp}) to improve test results on color images, despite our model being trained on gray-scale images. An example is shown in Fig.~\ref{fig:result_color}. The hybrid method shares the advantages of our original model that effectively preserves edges and textures and the cross-channel prior that reduces color artifacts.

\begin{figure*}
\centering
\subfigure[Ground truth]
{\includegraphics[width=.4\columnwidth]{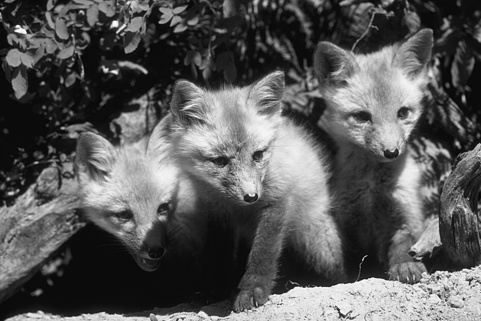}}
\subfigure[Noisy input (20.18dB)]
{\includegraphics[width=.4\columnwidth]{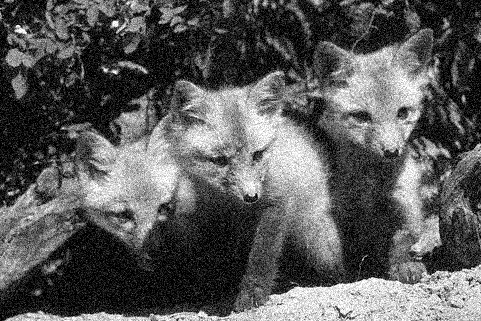}}
\subfigure[Iter 1 (22.85dB)]
{\includegraphics[width=.4\columnwidth]{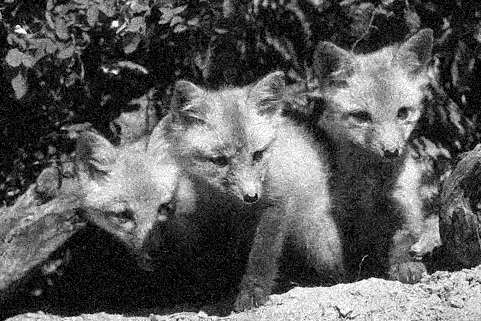}}
\subfigure[Iter 2 (25.93dB)]
{\includegraphics[width=.4\columnwidth]{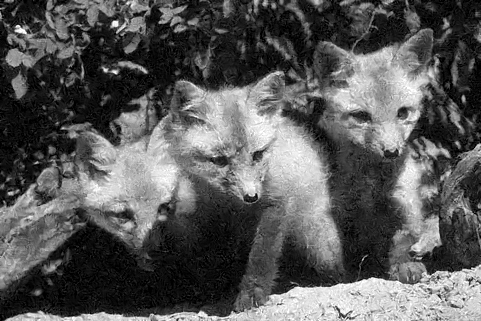}}
\subfigure[Iter 3 (28.14dB)]
{\includegraphics[width=.4\columnwidth]{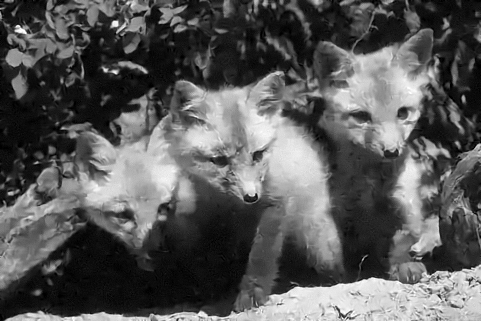}}
\caption{Results at each HQS iteration of our method on image denoising with noise level $\sigma=25$. Inside brackets show the PSNR values.}
\label{fig:result_denoise_intermediate}
\end{figure*}
\begin{figure*}
\vspace{-0.1in}
\centering
\subfigure[Ground truth]
{\includegraphics[width=.4\columnwidth]{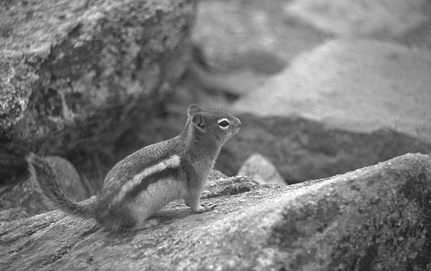}}
\subfigure[Blurry input (23.37dB)]
{\includegraphics[width=.4\columnwidth]{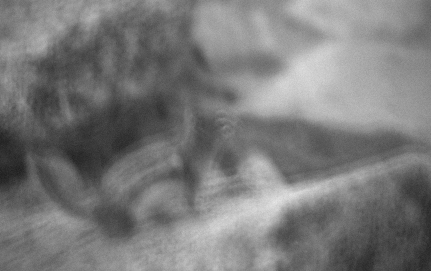}}
\subfigure[Iter 1 (27.32dB)]
{\includegraphics[width=.4\columnwidth]{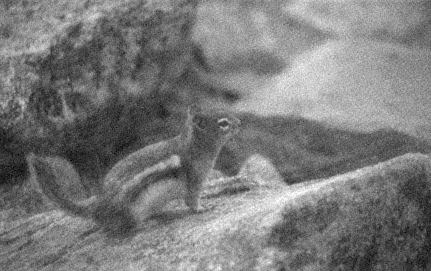}}
\subfigure[Iter 2 (28.48dB)]
{\includegraphics[width=.4\columnwidth]{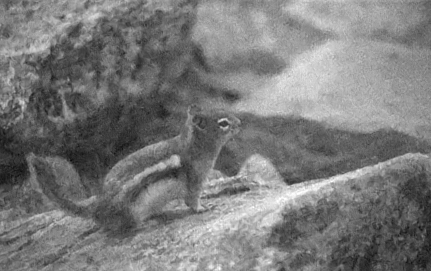}}
\subfigure[Iter 3 (29.36dB)]
{\includegraphics[width=.4\columnwidth]{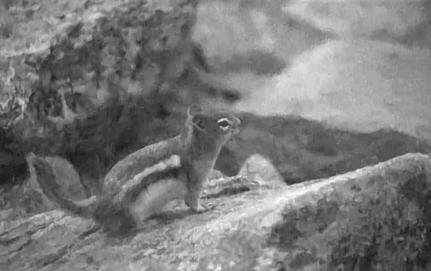}}
\caption{Results at each HQS iteration of our method on non-blind deconvolution with a 25$\times$25 PSF and noise level $\sigma=3$. }
\label{fig:result_deblur_intermediate}
\vspace{-0.2in}
\end{figure*}

\begin{figure}
\centering
\subfigure[Input]
{\includegraphics[width=.48\columnwidth]{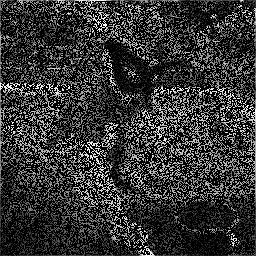}}
\subfigure[\small{Delaunay interp.(23.19dB)}]
{\includegraphics[width=.48\columnwidth]{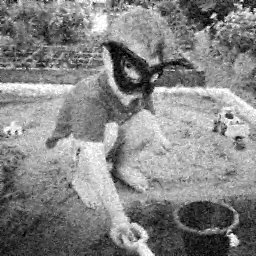}}\\
\vspace{-0.1in}
\subfigure[DTL$^5_3$ (25.10dB)]
{\includegraphics[width=.48\columnwidth]{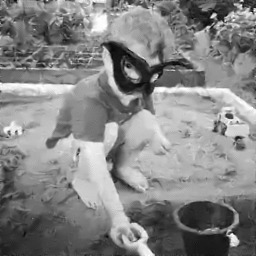}}
\subfigure[Ground truth]
{\includegraphics[width=.48\columnwidth]{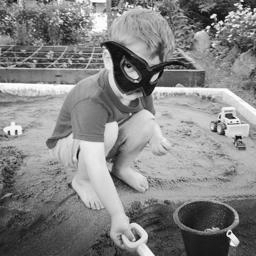}}
\caption{Experiment on joint denoising and inpainting task. The input image (a) misses 60\% pixels, and is corrupted with noise $\sigma=15$. Our method takes the result of Delaunay interpolation (b) as the initial estimation $\latentV^0$. Please zoom in for better view.}
\label{fig:result_inpaint_denoise}
\end{figure}

\smallskip
\noindent{\bf{Transferability to unseen tasks.}} Our method allows for new data-fidelity terms that are not contained in training, with no need for re-training. We demonstrate this flexibility with an experiment on the joint denoising and inpainting task shown in Fig.~\ref{fig:result_inpaint_denoise}. In this experiment, 60\% pixels of the input image are missing, and the measured 40\% pixels are corrupted with Gaussian noise with $\sigma=15$. Let vector $\V{a}$ be the binary mask for measured pixels. The sensing matrix $\sensM$ in Eq.~\ref{eq:FoE_model}, assumed to be known, is a binary diagonal matrix (hence $\sensM=\sensM\transpose=\sensM\transpose\sensM$) with diagonal elements $\V{a}$. To reuse our model trained on denoising/deconvolution tasks, we only need to specify $\sensM$ and $\weightFidelity$. The subproblems of our HQS framework are given in Eq.~\ref{eq:joint_denoise_inpaint_iter}.
\begin{eqnarray}
\begin{aligned}
&\slackPrimaryV^t = \prox_{\Theta} (\latentV^{t-1}),\\
&\latentV^t = (\weightFidelity \sensM\transpose\measV + \rho^t\slackPrimaryV^t)/(\weightFidelity \V{a}+\rho^t)
\label{eq:joint_denoise_inpaint_iter}
\end{aligned}
\end{eqnarray}

\smallskip
\noindent{\bf{Analysis of convergence and model complexity.}} To better understand the convergence of our method, in Fig.~\ref{fig:result_denoise_intermediate} and~\ref{fig:result_deblur_intermediate} we show the results of each HQS iteration of our method on denoising and non-blind deconvolution.

To understand the effect of model complexity and the number of HQS iteration on results, in Table~\ref{tab:result_model_complexity} we report test results of our method using models trained with different HQS iterations ($T$ in Algorithm~\ref{alg:alg_main}), and with different stages in $\prox_\Theta$ ($K$ in Algorithm~\ref{alg:alg_main}).

\begin{table}[h]
\centering
\caption{Test with different HQS iterations ($T$) and model stages ($K$) for image denoising. Average PSNR (dB) results on 68 images from~\cite{levin2011efficient} with noise $\sigma=15$ and $25$ are reported (before and after ``/'' in each cell respectively).}
\label{tab:result_model_complexity}
\begin{tabular}{cc|ccc}
\hline
                  &   & \multicolumn{3}{c}{\# HQS iterations} \\
                  &   & 1              & 3              & 5              \\ \hline
\multirow{3}{*}{\rotatebox[origin=c]{270}{\# stages}} & 1 & 29.80 / 26.81    & 30.89 / 28.12    & 30.96 / 28.28    \\
                  & 3 & 30.54 / 27.82    & 30.91 / 28.19    & 31.00 / 28.42    \\
                  & 5 & 30.54 / 27.83    & 30.92 / 28.18    & -              \\ \hline
\end{tabular}
\end{table}


\section{Conclusion}
\label{sec:conclusion}

In this paper, we proposed the discriminative transfer learning
framework for general image restoration. By combining advanced
proximal optimization algorithms and discriminative learning
techniques, a single training pass leads to a transferable model
useful for a variety of image restoration tasks and problem
conditions. Furthermore, our method is flexible and can be combined
with existing priors and likelihood terms after being trained,
allowing us to improve image quality on a task at hand. In spite of
this generality, our method achieves comparable run-time efficiency as
previous discriminative approaches, making it suitable for
high-resolution image restoration and mobile vision applications.

We believe that in future work, our framework incorporating advanced 
optimization with discriminative learning techniques can be extended 
to deep learning, for training more compact and shareable models, 
and to solve high-level vision problems.


{\small
\bibliographystyle{ieee}
\bibliography{egbib}
}

\end{document}